\documentclass[reqno]{amsart}
\usepackage{mathtools}

\usepackage{tikz}
\usetikzlibrary{arrows.meta}

\def\tr{{\rm Tr}}

\def\N{{\mathbb N}}

\def\R{{\mathbb R}}
\def\Z{{\mathbb Z}}

\def\unu{{\underline{\nu}}}

\def\utau{{\underline{\tau}}}

\def\cC{{\mathcal C}}
\def\cD{{\mathcal D}}

\def\cI{{\mathcal I}}

\def\cL{{\mathcal L}}
\def\cM{{\mathcal M}}

\def\cS{{\mathcal S}}
\def\cU{{\mathcal U}}

\def\1{{\bf 1}}
 
\def\uN{\underline{N}}

\def\Z{\theta}
\def\uZ{\underline{\Z}}

\def\diag{{\rm diag}}

\def\layersp{{\mathfrak L}}

\def\supp{{\rm supp}}

\def\CostN{\cC_{\uN}}
\def\CostNin{\CostN}
\def\CostNout{\widetilde\CostN}
\def\spec{{\rm spec}}
\def\tb{{\widetilde b}}

\def\X0{X_0}

\def\eqnn{\begin{eqnarray*}}
\def\eeqnn{\end{eqnarray*}}
\def\eqn{\begin{eqnarray}}
\def\eeqn{\end{eqnarray}}

\def\prf{\begin{proof}}
\def\endprf{\end{proof}}

\theoremstyle{plain}
\newtheorem{theorem}{Theorem}[section]
\newtheorem{definition}[theorem]{Definition}
\newtheorem{proposition}[theorem]{Proposition}

\newtheorem{lemma}[theorem]{Lemma}
\newtheorem{corollary}[theorem]{Corollary}

\numberwithin{equation}{section}

\begin{document} 

\title[Effective gradient flow equations in DL]
{Derivation of effective gradient flow equations  and dynamical truncation of training data in Deep Learning}

\author{Thomas Chen}
\address[T. Chen]{Department of Mathematics, University of Texas at Austin, Austin TX 78712, USA}
\email{tc@math.utexas.edu}  

\begin{abstract}
We derive explicit equations governing the cumulative biases and weights in Deep Learning with ReLU activation function, based on gradient descent for the Euclidean loss in the input layer, and under the assumption that the weights are, in a precise sense, adapted to the coordinate system distinguished by the activations. We show that gradient descent corresponds to a dynamical process in the input layer, whereby clusters of data are progressively reduced in complexity ("truncated") at an exponential rate that increases with the number of data points that have already been truncated. We provide a detailed discussion of several types of solutions to the gradient flow equations. A main motivation for this work is to shed light on the interpretability question in supervised learning.  
\end{abstract}

\maketitle

\tableofcontents

\section{Introduction}


In this paper, we continue our investigation of the interpretability (or "black box") problem in Deep Learning in \cite{ch-6,cheewa-1,cheewa-3,cheewa-4,cheewa-5}, specifically in the context of supervised learning. 
Our aim in the work at hand is to elucidate the training dynamics as a geometrically intuitive process in the input layer. To this end, we determine the action of the gradient flow in terms of a dynamical system acting on the training data in the input layer, by deriving the effective equations for the cumulative weights and biases (for simplicity of presentation, we choose the ReLU activation function). We analyze several classes of solutions, and show that the gradient flow in DL is equivalent to the action of dynamical truncations in input space, by which clusters of training data are progressively reduced in their geometric complexity;   under the right circumstances, they are contracted to points. The latter corresponds to a dynamical realization of the phenomenon of neural collapse computationally evidenced in \cite{PHD20}, and leads to zero training loss. 

The purpose of this paper is to elucidate the geometry of training dynamics in the input layer, and we choose simplifications that eliminate complications that would cloud the core structures.
To this end, we consider a DL network with equal dimensions in all layers. Accordingly, we associate training vectors $x^{(0)}\in\R^Q$ with the input layer, and define hidden layers, indexed by $\ell=1,\dots,$, where recursively,
\eqn
	x^{(\ell)} = \sigma(W_\ell x^{(\ell-1)} + b_\ell) 
	\;\;\in\R^{Q} \,.
\eeqn
The map in the $\ell$-th layer is parametrized by the weight matrix $W_\ell\in\R^{Q\times Q}$ and bias vector $b_\ell\in\R^{Q}$. We choose the activation function $\sigma$ to be ReLU (the ramp function, $\sigma(x)=\max\{0,x\}$), and use the convention that its (weak) derivative is given by 
\eqn 
	\sigma'(x)=h(x)=
	\left\{
	\begin{array}{cc}
	1&x>0\\
	0&x\leq0
	\end{array}
	\right.
\eeqn
Both $\sigma$ and $h$ are defined to act component-wise on $x\in\R^Q$.

Assuming that all $W_\ell\in GL(Q)$ are invertible, we define the {\em cumulative parameters}
\eqn\label{eq-cumulWB-def-1-0-0}
	W^{(\ell)} &:=& W_\ell W_{\ell-1}\cdots W_1
	\nonumber\\
	b^{(\ell)} &:=& W_\ell\cdots W_2 b_1 
	+ \cdots + W_2 b_{\ell-1} + b_\ell
	\nonumber\\
	\beta^{(\ell)}&:=& (W^{(\ell)})^{-1}b^{(\ell)}
	= \sum_{j=1}^\ell (W^{(j)})^{-1}b_{j}
\eeqn
for $\ell=1,\dots,L$, and
\eqn 
	\beta^{(L+1)}&:=& (W_{L+1})^{-1}\beta^{(L)} 
\eeqn 
in the output layer. Introducing the affine maps
\eqn\label{eq-affine-def-1-0-0-0}
	a^{(\ell)}(x) := W^{(\ell)}x+b^{(\ell)}
\eeqn  
we define the {\em truncation maps}
\eqn 
	\tau^{(\ell)} (x) &:=& 
	(a^{(\ell)})^{-1}\circ\sigma\circ a^{(\ell)}(x)
	\nonumber\\
	&=&
	(W^{(\ell)})^{-1}(\sigma(W^{(\ell)}x+b^{(\ell)}) -b^{(\ell)})
	\nonumber\\
	&=& 
	(W^{(\ell)})^{-1}\sigma(W^{(\ell)}(x+\beta^{(\ell)})) 
	-\beta^{(\ell)} \,,
\eeqn 
in the same way as in \cite{cheewa-2,cheewa-4}. The $\ell$-th truncation maps is the pullback of the activation map under $a^{(\ell)}$; that is, $a^{(\ell)}$ maps a vector $x$ in input space to the $\ell$-th layer where $\sigma$ acts on it, and subsequently, $(a^{(\ell)})^{-1}$ maps the resulting vector back to the input layer. Accordingly, the gradient flow of cumulative weights and biases induces a dynamics of time-dependent truncation maps acting on the training data in the input layer. 

We assume that the reference outputs (labels) are given by $y_{\ell}\in\R^Q$, $\ell=1,\dots,Q$, and denote the training inputs belonging to the label $y_\ell$ by $x_{\ell,i}^{(0)}\in \R^{Q}$, $\ell=1,\dots,Q$, $i=1,\dots,N_\ell$. We will write $\uN:=(N_1,\dots,N_Q)$. Moreover, we assume $L=Q$ for the number of hidden layers.

As we will explain in detail in Section \ref{sec-defmodel-1-0-0}, the standard $\cL^2$ loss 
\eqn\label{eq-CostNpullb-def-1-0-0}
	\CostNout = 
	\frac12\sum_{j=1}^Q \frac1{N_j} \sum_{i=1}^{N_j}|W^{(L+1)}
	\big(\underline\tau^{(L)}(x_{j,i}^{(0)})
	-(W^{(L+1)})^{-1}y_j\big)|^2_{\R^{Q}} 
\eeqn 
is defined with the pullback metric in input space with respect to the map $W^{(L+1)}$ from input to output space. In gradient descent algorithms, $W_{L+1}$ (and thus, $W^{(L+1)}$) are often treated as dynamical parameters, and the non-Euclidean, time dependent metric introduces many of the known complications ("loss landscape").

Here, we propose to investigate the Euclidean $\cL^2$ loss in the input space,
\eqn\label{eq-CostNEucl-def-1-0-0}
	\CostNin := 
	\frac12\sum_{j=1}^Q \frac1{N_j} \sum_{i=1}^{N_j}| 
	\underline\tau^{(L)}(x_{j,i}^{(0)})
	-(W^{(L+1)})^{-1}y_j|^2_{\R^{Q}}
\eeqn 
where we study the gradient flow at fixed $W^{(L+1)}$.

Moreover, we note that the choice of the activation map $\sigma$ distinguishes a specific coordinate system. The polar decomposition of the cumulative weight yields
\eqn 
	W^{(\ell)} = |W^{(\ell)}| R_\ell 
\eeqn 
where $R_\ell = |W^{(\ell)}|^{-1} W^{(\ell)} \in O(Q)$ is an orthogonal matrix, and $|W^{(\ell)}|$ is symmetric. Accordingly,
\eqn
	|W^{(\ell)}| 
	= \widetilde R_\ell^T W^{(\ell)}_* \widetilde R_\ell
\eeqn
where $W^{(\ell)}_*$ is diagonal, and $\widetilde R_\ell\in SO(Q)$ accounts for the degree of freedom of rotating the coordinate system in which $\sigma$ is defined.
In this paper, we choose the cumulative weights to be {\em adapted to the activation} in that $\widetilde R_\ell=\1$, so that 
\eqn 
	W^{(\ell)} = W^{(\ell)}_* R_\ell 
\eeqn
with $W^{(\ell)}_*\geq0$ diagonal. One then observes that the truncation maps become independent of $W^{(\ell)}_*$ (a consequence of $(W^{(\ell)}_*)^{-1}\sigma(W^{(\ell)}_* x)=\sigma(x)$), and that therefore, $\beta^{(\ell)}\in\R^Q$ and $R_\ell\in O(Q)$ parametrize the DL network. The analysis of more general situations including variable layer dimensions and general weights with $\widetilde R_\ell\neq\1$ are left for future work.

We denote the empirical probability distribution on $\R^Q$, associated to the $\ell$-th cluster of training inputs, by
\eqn\label{eq-emp-probmeas-1-0-0}
	\mu_\ell(x):= \frac1{N_\ell}\sum_{i=1}^{N_\ell} 
	\delta(x-x_{\ell,i}^{(0)}) \,,
\eeqn
where $\delta$ is the Dirac delta distribution. We write
\eqn
	\widetilde y_\ell := (W^{(Q+1)})^{-1}y_\ell 
\eeqn
for notational convenience, with $W^{(Q+1)}$ fixed.

We define the notion of {\em cluster separated truncations} that accounts for the $\ell$-th truncation map acting nontrivially only on training inputs in the $\ell$-th cluster, but acting on all other clusters as the identity, $\tau^{(\ell)}(x_{\ell',i}^{(0)})=x_{\ell',i}^{(0)}$ for all $\ell'\neq\ell$. This property was crucially used in \cite{cheewa-2} and \cite{cheewa-4}. It requires the supports of $\mu_\ell$, $\ell=1,\dots,Q$,  to be sufficiently separated from one another. See for instance \cite{phulam20} for algorithms detecting the clustering of data.

\subsection{Summary of main results}

We then prove, in Theorem \ref{thm-gradflow-input-1-0}, that for cluster separated truncations, the gradient flow for the cumulative weights and biases is given by the effective equations
\eqn\label{eq-gradflow-moments-1-0-0-1}
	\partial_s(\beta^{(\ell)}+\widetilde y_\ell)
	&=&  
	 - R_\ell^T J_0^{(\ell) \perp} R_\ell
	( \beta^{(\ell)}+\widetilde y_\ell) \,
	\nonumber\\ 
	\partial_s R_\ell 
	&=& - \, \Omega_\ell  R_\ell
\eeqn 
where
\eqn
	J_0^{(\ell) \perp} = \int_{
	\R^Q\setminus\R^Q_+}
	dx \, \mu_\ell(a_{R_\ell,\beta^{(\ell)}}^{-1}(x)) \, H^\perp(x) \,
\eeqn
is a diagonal matrix with 
\eqn
	H^\perp(x)&=&\1_{Q\times Q}-H(x)
	\nonumber\\
	H(x)&=&\diag(h(x_i)\,;\,i=1,\dots,Q)
\eeqn
and
\eqn\label{eq-Om-commut-def-1-0-0}
	\Omega_\ell
	=
	\int_{\R^Q\setminus(\R_+^Q\cup \R_-^Q)}  
	dx \, \mu_\ell(a_{R_\ell,\beta^{(\ell)}}^{-1}(x))
	\big[H(x)\;,\;
	M^{(\ell)}(x) \big] \,,
\eeqn
where $[A,B]=AB-BA$ is the commutator of $A,B\in\R^{Q\times Q}$, and
\eqn
	M^{(\ell)}(x) :=  
	\frac12\Big(x (\beta^{(\ell)}+\widetilde y_\ell)^T R_\ell^T 
	+ R_\ell(\beta^{(\ell)}+\widetilde y_\ell) x^T\Big)
	\,.
\eeqn
In Proposition \ref{prop-equi-0-1} and Theorem \ref{eq-gradflow-conv-1-0-0}, we prove that 
\begin{itemize}
\item
The pair $(\beta^{(\ell)},R_\ell)$ is an equilibrium solution if 
\eqn 
	\supp\Big(\mu_\ell \circ a_{R_\ell,\beta^{(\ell)}}^{-1}\Big)
	\subset \R_+^Q
\eeqn 
or 
\eqn 
	\supp\Big(\mu_\ell \circ a_{R_\ell,\beta^{(\ell)}}^{-1}\Big)
	\subset \R_-^Q \,.
\eeqn 
In the first case, $\tau^{(\ell)}$ acts as the identity on the $\ell$-th cluster, while in the second case, the $\ell$-th cluster is contracted to a point.
\item
If the initial data $(\beta^{(\ell)}(0),R_\ell(0))$ is such that 
\eqn
	\supp\Big(\mu_\ell \circ 
	a_{R_\ell(0),\beta^{(\ell)}(0)}^{-1}\Big)
	\cap \R^Q\setminus(\R_+^Q\cup\R_-^Q) \neq \emptyset \,,
\eeqn
and the support of $\mu_\ell \circ a_{R_\ell,\beta^{(\ell)}}^{-1}$ is suitably geometrically positioned, and sufficiently concentrated in $\R_-^Q$ in a manner that
\eqn 
	J_0^{(\ell) \perp} > 1-\eta
\eeqn
for a small constant $\eta$, the following holds. 

The solution of the gradient flow translates $\mu_\ell \circ a_{R_\ell(s),\beta^{(\ell)}(s)}^{-1}$ into $\R_-^Q$ in finite time $s=s_1<\infty$, and $\beta^{(\ell)}(s)\rightarrow -\widetilde y_\ell$ converges exponentially as $s\rightarrow\infty$. For $s>s_1$, the weight matrix $R_\ell(s)=R_\ell(s_1)$ is stationary. In particular, this implies that the entire $\ell$-th cluster is collapsed into the point $\beta^{(\ell)}(s)$ for $s>s_1$. See Figure \ref{fig-1} below.

This provides an interpretation of the phenomenon of neural collapse on the level of training data in input space, as computationally evidenced in \cite{PHD20}. See also \cite{cheewa-5,mixparpi22,wwoj}.
\end{itemize}

In Section \ref{ssec-beta-ex-flow-1-0}, we present a detailed analysis of the dynamics of the cumulative bias $\beta^{(\ell)}(s)$ at fixed $R_\ell$, and show that it converges exponentially to $-\widetilde y_\ell$, at a rate that increases with every additional training input that is truncated.

In Theorem \ref{thm-gradflow-general-1-0}, we derive the gradient flow equations in the general case, without the assumption of cluster separated truncations. They describe dynamical truncations of clusters that are renormalized by the intersection of the positive sectors of all truncation maps. The geometry of the resulting configurations of data is significantly more complicated than the case discussed above, see for instance \cite{grilinmeywu}. An analysis of the dynamics is left for future work. 

In Sections \ref{sec-std-gradflow-1-0-0} and \ref{sec-std-clustered-1-0-0}, we address two situations in which the gradient flow for the standard loss \eqref{eq-CostNpullb-def-1-0-0} can be explicitly controlled. In Proposition \ref{prp-conservation-1}, we prove, for fully collapsed initial data, that there exist matrix valued integrals of motion providing a spectral gap that drives the loss to exponentially converge to zero.

We remark that the situations considered in this work cover {\em underparametrized} DL networks as in \cite{cheewa-2,cheewa-4}. The constructive determination of loss minimizers in overparametrized networks, and geometric generalization bounds are addressed in \cite{chechiewamoo-1,ch-6,cheewa-5}; for related works, see for instance \cite{GVS15,KMTM24,tia17,WGL20} and the references therein.

\begin{figure}
\label{fig-1}

\begin{tikzpicture}[
    x=1cm,y=1cm,
    >=Latex,
    axis/.style={black, line width=0.6pt},
    traj/.style={blue!70!black, line width=0.9pt, -{Latex[length=2mm]}},
    pt/.style={circle, fill=blue!70!black, inner sep=1.5pt}
]

\newcommand{\panelaxes}{
  \draw[axis] (-3.8,0) -- (1.5,0);
  \draw[axis] (0,-3.4) -- (0,1.2);
}

\def\blobpath{
  (0, 0.8)
  .. controls (0.6, 0.8) and (1.2, 0.4) .. (1.2, -0.2)
  .. controls (1.2, -0.8) and (0.6, -1.4) .. (-0.2, -1.4)
  .. controls (-1.0, -1.4) and (-1.4, -0.6) .. (-1.4, 0)
  .. controls (-1.4, 0.8) and (-0.6, 0.8) .. cycle
}

\newcommand{\drawshiftedblob}[2]{
  \begin{scope}
    \clip (-5,-5) rectangle (0,0);
    \fill[gray!20, fill opacity=0.6, shift={(#1,#2)}] \blobpath;
  \end{scope}
  
  \begin{scope}
    \clip (0,0) -- (0,-5) -- (5,-5) -- (5,5) -- (-5,5) -- (-5,0) -- cycle;
    \fill[gray!70, fill opacity=0.7, shift={(#1,#2)}] \blobpath;
  \end{scope}
  
   \begin{scope}
    \clip (5,5) rectangle (0,0);
    \fill[gray!100, fill opacity=0.7, shift={(#1,#2)}] \blobpath;
  \end{scope}
  
  \draw[draw=black, line width=0.7pt, shift={(#1,#2)}] \blobpath;
}

\begin{scope}[shift={(0,0)}]
  \node[anchor=west] at (-3.8,1.5) {$s=0$};
  
  \panelaxes
  \drawshiftedblob{0}{0}
  
  \draw[blue!70!black, line width=1.6pt] (0,0) -- (0, 0.8);
  \draw[traj, blue!70!black, thick, densely dashed] (-0.9, 0.5) -- (0, 0.5);
  \draw[traj, blue!70!black, thick, densely dashed] (-0.9, 0.2) -- (0, 0.2);
  \draw[blue!70!black, line width=1.6pt] (0,0) -- (1.2, 0);
  \draw[traj, blue!70!black, thick, densely dashed] (0.3, -0.8) -- (0.3, 0);
  \draw[traj, blue!70!black, thick, densely dashed] (0.7, -0.8) -- (0.7, 0);
\end{scope}

\begin{scope}[shift={(6,0)}]
  \node[anchor=west] at (-3.8,1.5) {$s<s_1$};
  
  \panelaxes
  \drawshiftedblob{-0.9}{-0.6}

  \draw[blue!70!black, line width=1.6pt] (0,0) -- (0, 0.2);
  \draw[traj, blue!70!black, thick, densely dashed] (-0.9, 0.2) -- (0, 0.2);
  \draw[blue!70!black, line width=1.6pt] (0,0) -- (0.3, 0);
  \draw[traj, blue!70!black, thick, densely dashed] (0.3, -0.8) -- (0.3, 0);
\end{scope}

\begin{scope}[shift={(0,-6.5)}]
  \node[anchor=west] at (-3.8,1.5) {$s=s_1$};
  
  \panelaxes
  \drawshiftedblob{-1.2}{-0.8}

  \draw[traj, thick, densely dashed] (-1.2, 0) -- (0,0);
  \draw[traj, thick, densely dashed] (0, -1.0) -- (0,0);

  \node[pt] at (0, -1.0) {};   
  \node[pt] at (-1.2, 0) {};   
  
  \node[pt, opacity=0.8] at (0,0) {};
\end{scope}

\begin{scope}[shift={(6,-6.5)}]
  \node[anchor=west] at (-3.8,1.5) {$s>s_1$};
  
  \panelaxes
  \drawshiftedblob{-1.8}{-1.4}

  \draw[traj, thick, densely dashed] (-2.75, -0.7) -- (0,0); 
  \draw[traj, thick, densely dashed] (-0.7, -2.05) -- (0,0); 
  
  \node[pt, opacity=0.8] at (0,0) {};
\end{scope}

\end{tikzpicture}

\caption{At the initial time $s=0$, the training data distribution intersects the complement of the negative sector of the ReLU activation, in its co-moving frame, and moves deeper into it as $s>0$ elapses. At the time $s=s_1$, all of it lies entirely in its negative sector, and is mapped to the origin. As $s>s_1$, it moves further into the negative sector, until the bias vector asymptotically matches the preimage of the label, $\beta^{(\ell)}(s)+\widetilde y_\ell\rightarrow0$ as $s\rightarrow\infty$. In the pictures above, the light gray region of the data distribution is mapped to the origin, the medium gray regions to the adjacent positive coordinate axis (blue line segments), and the dark gray region is unaffected by ReLU.}

\end{figure}


\section{Definition of the mathematical model}
\label{sec-defmodel-1-0-0}

We consider the setting of supervised learning in a deep network with $L$ hidden layers. We associate the space $\layersp_0=\R^{M_0}$ to the input layer, the spaces $\layersp_\ell = \R^{M_\ell}$ to the hidden layers, $\ell=1,\dots,L$, and $\layersp_{L+1}=\R^M_{L+1}$ to the output layer, with $M_0,\dots,M_{L+1}\in\N$.

We will specifically assume that the reference outputs (labels) are given by $y_{\ell}\in\R^Q$, $\ell=1,\dots,Q$, so that $M_{L+1}=Q$. We denote the training inputs belonging to the label $y_\ell$ by $x_{\ell,i}^{(0)}\in \layersp_0=\R^{M_0}$, $\ell=1,\dots,Q$, $i=1,\dots,N_\ell$. 

We will refer to $\{x_{\ell,i}^{(0)}\}_{i=1}^{N_\ell}\subset \R^Q$ as the {\em $\ell$-th cluster of training inputs}, and will use the multiindex notation $\uN:=(N_1,\dots,N_Q)\in\N_0^Q$, with $N:=\sum_{j=1}^Q N_j$.

The $\ell$-th layer, defined on $\layersp_\ell=\R^{M_\ell}$, recursively determines the map
\eqn
	x_j^{(\ell)} = \sigma(W_\ell x_j^{(\ell-1)} + b_\ell) \;\;\in\layersp_\ell =\R^{M_\ell}
\eeqn
parametrized by the weight matrix $W_\ell\in\R^{M_\ell\times M_{\ell-1}}$ and bias vector $b_\ell\in\R^{M_\ell}$. We choose the activation function $\sigma$ to be the same for every $\ell$. 
Accordingly, we define the $\ell$-th layer cluster averages
\eqn 
	\overline{x_j}^{(\ell)} := 
	\frac1{N_j}\sum_{i=1}^{N_j}x_{j,i}^{(\ell)}
\eeqn 
and deviations
\eqn 
	\Delta x_{j,i}^{(\ell)} := 
	x_{j,i}^{(\ell)} - \overline{x_j}^{(\ell)} 
\eeqn 
for $j=1,\dots,Q$.
The output layer is associated with the map 
\eqn
	x_j^{(L+1)} = W_{L+1} x_j^{(L)} + b_{L+1} \;\;\in\layersp_{L+1}=\R^{Q} \,,
\eeqn
and includes no activation function.
We assume that $M_\ell\leq M_{\ell-1}$ are non-increasing.

We denote the vector of parameters by 
\eqn\label{eq-parmsp-K-def-1-0-0}
	\uZ \in\R^K
	\;\;\;,\;\;\;
	K = \sum_{\ell=1}^{L+1} (M_\ell M_{\ell-1}+M_\ell )
\eeqn
containing the components of all weights $W_\ell$ and biases $b_\ell$, $\ell=1,\dots,L+1$, including those in the output layer.  

To begin with, we consider the case $M_\ell=Q$, $\ell=1,\dots,L$, in which the dimensions of the input and hidden layer spaces are all $Q$.

Assuming that all $W_\ell\in GL(Q)$, we define the {\em cumulative parameters}
\eqn\label{eq-cumulWB-def-1-0}
	W^{(\ell)} &:=& W_\ell W_{\ell-1}\cdots W_1
	\nonumber\\
	b^{(\ell)} &:=& W_\ell\cdots W_2 b_1 + \cdots + W_2 b_{\ell-1} + b_\ell
	\nonumber\\
	\beta^{(\ell)}&:=& (W^{(\ell)})^{-1}b^{(\ell)}
	= \sum_{j=1}^\ell (W^{(j)})^{-1}b_{j}
\eeqn
for $\ell=1,\dots,L$, and
\eqn 
	\beta^{(L+1)}&:=& (W_{L+1})^{-1}\beta^{(L)} 
\eeqn 
in the output layer. Introducing the affine maps
\eqn\label{eq-affine-def-1-0-0}
	a^{(\ell)}(x) := W^{(\ell)}x+b^{(\ell)}
\eeqn  
we define the {\em truncation maps}
\eqn 
	\tau^{(\ell)} (x) &:=& 
	(a^{(\ell)})^{-1}\circ\sigma\circ a^{(\ell)}(x)
	\nonumber\\
	&=&
	(W^{(\ell)})^{-1}(\sigma(W^{(\ell)}x+b^{(\ell)}) -b^{(\ell)})
	\nonumber\\
	&=& 
	(W^{(\ell)})^{-1}\sigma(W^{(\ell)}(x+\beta^{(\ell)})) -\beta^{(\ell)} \,.
\eeqn 
We note that
\eqn 
	a^{(\ell)}: \layersp_0 \longrightarrow \layersp_\ell \,,
\eeqn 
and
\eqn
	\tau^{(\ell)}: \layersp_0\xrightarrow[]{a^{(\ell)}}\layersp_\ell 
	\xrightarrow[]{\sigma}\layersp_\ell 
	\xrightarrow[]{(a^{(\ell)})^{-1}}\layersp_0 \,.
\eeqn 
That is, the vector $x\in\layersp_0$ in the input layer is mapped to the $\ell$-th layer via $a^{(\ell)}$ where the activation function $\sigma$ acts on its image, and is subsequently pulled back to the input space via $(a^{(\ell)})^{-1}$. 

\begin{definition}
	We denote the sets $\cS_{\ell}^+$, $\cS_{\ell}^-$, defined by
\eqn\label{eq-cSell-def-1-0-0}
	\cS_{\ell}^+ &:=&\{x\in\R^Q \, | \, 
	a^{(\ell)}(x)  \in\R_+^Q \}
	\nonumber\\
	\cS_{\ell}^- &:=&\{x\in\R^Q \, | \, 
	a^{(\ell)}(x)  \in\R_-^Q \}
\eeqn 
as the {\em positive}, respectively, {\em negative sector} of the truncation map $\tau^{(\ell)}$, and
\eqn 
	\cS_\ell^\perp:=\R^Q\setminus\cS_\ell^+ \,.
\eeqn  
We say that $x\in\layersp_0\cong\R^Q$ is 
\begin{itemize}
	\item
	untruncated by $\tau^{(\ell)}$ if $x\in\cS_{\ell}^+$,
	\item
	partially truncated by $\tau^{(\ell)}$ if $x\in\cS_\ell^\perp$,  
	\item
	fully truncated by $\tau^{(\ell)}$ if $x\in \cS_{\ell}^-$, and
	\item
	truncated in the $r$-th coordinate direction if $(W_\ell(x+\beta^{(\ell)}))_r\in\R_-$.
\end{itemize}
Moreover, we say that a set $\{x_i\}_i$ is fully truncated or   untruncated if all $x_i$ are fully truncated, respectively untruncated. Otherwise, we say that $\{x_i\}_i$ is partially truncated.
\end{definition}

Clearly, if $x$ is untruncated, then
\eqn
	\tau^{(\ell)}(x) = x 
	\;\;\;\Leftrightarrow\;\;\; 
	x\in\cS_{\ell}^+ \,,
\eeqn 
that is, $\cS_{\ell}^+\subset\R^Q$ is the fixed point set of $\tau^{(\ell)}$.
On the other hand, if $x$ is fully truncated,
\eqn
	\tau^{(\ell)}(x) = -\beta^{(\ell)}
	\;\;\;\Leftrightarrow\;\;\; 
	x\in\cS_{\ell}^- \,.
\eeqn 
Thus in particular, if $x\in \cS_{\ell}^+\cup \cS_{\ell}^-$, it follows that $\tau^{(\ell)}(x)$ is independent of $W^{(\ell)}$, and if $x\in \cS_{\ell}^+$, then $\tau^{(\ell)}(x)$ is also independent of $\beta^{(\ell)}$.

Then, defining
\eqn 
	\underline\tau^{(\ell)}:= \tau^{(\ell)}\circ\cdots\circ\tau^{(1)} \,,
\eeqn 
the vectors in the $\ell$-th hidden layer are given by
\eqn 
	x_{j,i}^{(\ell)} = W^{(\ell)} \underline\tau^{(\ell)}(x_{j,i}^{(0)}) \,,
\eeqn 	
for $\ell=1,\dots,L$.

The vectors in the output layer are obtained by
\eqn 
	x_{j,i}^{(L+1)} 
	= W_L   x_{j,i}^{(L)} 
	= W^{(L+1)}\underline\tau^{(L)}(x_{j,i}^{(0)}) \,,
\eeqn 
for all $j=1,\dots,L$, and $i=1,\dots,N_j$. 
That is, the vectors $\tau^{(\ell)}(x_{j,i}^{(0)})$ in the input layer are mapped by $W^{(L+1)}=W_{L+1}W_L\cdots W_1$, via
\eqn
	W^{(L+1)} 
	: \layersp_0 \xrightarrow[]{W_1}
	\layersp_1\xrightarrow[]{W_2}\cdots\xrightarrow[]{W_{L+1}}\layersp_{L+1} \,,
\eeqn 
to the output layer. We will assume that $W^{(L+1)}$ has full rank.

In the input layer, there are two natural metrics associated with this problem. The Euclidean metric on $\layersp_0$ on one hand,
\eqn
	|x|_{\layersp_0} = |x|_{\R^{M_0}} \,,
\eeqn 
and on the other hand, the pullback metric under $W^{(L+1)}: \layersp_0 \rightarrow\layersp_{L+1}$ obtained from the Euclidean metric on $\layersp_{L+1}$, 
\eqn\label{eq-pullb-L0-def-1-0}
	|x|_{\layersp_0, W^{(L+1)}} := |W^{(L+1)} x|_{\R^{M_{L+1}}}
\eeqn 
with metric tensor $(W^{(L+1)} )^TW^{(L+1)}:\layersp_0\rightarrow\layersp_0$.

\subsection{Standard loss is pullback loss in input layer}
\label{ssec-standcost-1-0-0}
The standard $\cL^2$ loss function is given by
\eqn\label{eq-CostNpullb-def-1-0}
	\CostNout &=& \frac12\sum_{j=1}^Q \frac1{N_j} \sum_{i=1}^{N_j}
	|x_{j,i}^{(L+1)}-y_j|^2_{\R^{M_{L+1}}}
	\nonumber\\
	&=&\frac12\sum_{j=1}^Q \frac1{N_j} \sum_{i=1}^{N_j}|W_{L+1}\big(x_{j,i}^{(L)}-W_{L+1}^{-1}y_j\big)|^2_{\R^{M_{L+1}}}
	\nonumber\\
	&=&\frac12\sum_{j=1}^Q \frac1{N_j} \sum_{i=1}^{N_j}|W^{(L+1)}
	\big(\underline\tau^{(L)}(x_{j,i}^{(0)})
	-(W^{(L+1)})^{-1}y_j\big)|^2_{\R^{M_{L+1}}}
	\nonumber\\
	&=&\frac12\sum_{j=1}^Q \frac1{N_j} \sum_{i=1}^{N_j}| 
	\underline\tau^{(L)}(x_{j,i}^{(0)})
	-(W^{(L+1)})^{-1}y_j|^2_{\layersp_0, W^{(L+1)}}
\eeqn 
That is, the standard loss is defined by use of the pullback metric \eqref{eq-pullb-L0-def-1-0} in $\layersp_0$ under $W^{(L+1)}: \layersp_0 \rightarrow\layersp_{L+1}$ obtained from the Euclidean metric in the output space $\layersp_{L+1}$. For every $j=1,\dots,Q$, it measures, relative to the pullback metric, the distance of the points $\tau^{(L)}(x_{j,i}^{(0)})$, $i=1,\dots,N_j$, to the preimage of the reference vectors (labels), $(W^{(L+1)})^{-1}y_j$.

Training of the DL network corresponds to finding global, or at least sufficiently good local minimizers of the loss function. The predominant approach is to employ the gradient flow $\partial_s\uZ=-\nabla_{\uZ}\CostNout$ in parameter space \eqref{eq-parmsp-K-def-1-0-0}, see \cite{ch-6,cheewa-5} for a discussion of geometric aspects of this problem. See also \cite{kaw16,zoulia18}.

This choice of the loss function introduces the following challenges:
\begin{itemize}
	\item It defines a metric in input space with metric tensor $(W^{(L+1)})^T W^{(L+1)}$, which is itself a time dependent parameter under the gradient flow. 
	\item Because of
	\eqn 
		x_{j,i}^{(L)}=W^{(L)}\tau^{(L)}(x_{j,i}^{(0)}) \,,
	\eeqn 
	it follows that the definition of
	\eqn
		x_{j,i}^{(L+1)}=W_{L+1}W^{(L)}\tau^{(L)}(x_{j,i}^{(0)})
	\eeqn 
	exhibits the issue that since $W^{(L)}$ is unknown, multiplication with the unknown $W_{L+1}$ introduces a degeneracy; namely, the pullback metric in $\layersp_0$ is invariant under 
	\eqn\label{eq-WL-redund-1-0}
		W^{(L)}\rightarrow A W^{(L)}
		\;\;\;,\;\;\;
		W^{(L+1)}\rightarrow W^{(L+1)} A^{-1}
	\eeqn 
	for any $A\in GL(M_L)$. The relevance of including $W_{L+1}$ in the output space $\layersp_{L+1}$ is to match $x_{j,i}^{(L)}$ to the reference outputs $y_j$, but including it in the definition of the pullback metric introduces a redundance.
	\item 
	The presence of $W_{L+1}$ in this form unnecessarily complicates the geometric structure of the gradient flow, as we will see below.
\end{itemize} 

\subsection{Euclidean loss in input layer}
For the above reasons, our key objective in this paper is to study the gradient flow generated by the Euclidean loss function in input space,
\eqn\label{eq-CostNEucl-def-1-0}
	\CostNin &=& 
	\frac12\sum_{j=1}^Q \frac1{N_j} \sum_{i=1}^{N_j}| 
	\underline\tau^{(L)}(x_{j,i}^{(0)})
	-(W^{(L+1)})^{-1}y_j|^2_{\R^{M_0}}
\eeqn 
Notably, in \eqref{eq-CostNEucl-def-1-0}, the reference output vectors $y_j\in\layersp_{L+1}$ are pulled back to $\layersp_0$ via $(W^{(L+1)})^{-1}$.
Combined with weight matrices adapted to the activation function $\sigma$, we will elucidate the natural geometrical interpretation of the action of the gradient flow in input space. It turns out to be quite intuitive and simple; the geometric understanding thus obtained will open up the path to gradient descent algorithms that do not require backpropagation.

For simplicity of exposition, we will assume that the number of hidden layers is $L=Q$, and that all layers have the same dimension, $M_\ell=Q$, $\ell=1,\dots,Q$. The general case will be addressed in future work.

Instead of the parameters $(W_\ell,b_\ell)_\ell$ that are usually used for the gradient flow, we will instead study the gradient flow of the cumulative parameters $(W^{(\ell)},\beta^{(\ell)})_\ell$. For different values of $\ell,\ell'$, the cumulative weights and biases $(W^{(\ell)},\beta^{(\ell)})$ and $(W^{(\ell')},\beta^{(\ell')})$ are independent parameters. 

\subsection{Weights adapted to the activation}

It is important to note that the activation function (which we think of as ReLU or a smooth mollification of ReLU) singles out a distinguished coordinate system.
Namely, in the $\ell$-th layer, the definition of
\eqn\label{eq-layerell-basis-1-0} 
	\sigma:\layersp_\ell\rightarrow\layersp_\ell
	\;\;\;,\;\;\;
	(x_1,\dots,x_Q)^T\mapsto ((x_1)_+,\dots,(x_Q)_+)^T
\eeqn 
depends on the choice of the coordinate system.

By assumption, for $\ell=1,\dots,Q+1$, the cumulative weight matrix $W^{(\ell)}:\layersp_0\rightarrow\layersp_\ell$ is an element of $\R^{Q\times Q}$ and we assume it to be invertible. It admits the polar decompositon
\eqn 
	W^{(\ell)} = |W^{(\ell)}| R_\ell 
\eeqn 
where $R_\ell = |W^{(\ell)}|^{-1} W^{(\ell)} \in O(Q)$ is an orthogonal matrix. Since $|W^{(\ell)}|$ is symmetric, 
\eqn
	|W^{(\ell)}| = \widetilde R_\ell^T W^{(\ell)}_* \widetilde R_\ell
\eeqn
where $W^{(\ell)}_*$ is diagonal, and $\widetilde R_\ell\in SO(Q)$ maps the eigenbasis of $|W^{(\ell)}|$ to the orthonormal coordinate system distinguished by the definition of the activation map, \eqref{eq-layerell-basis-1-0}.

Given $x\in\layersp_0$, the map
\eqn 
	\sigma(W^{(\ell)} x) = 
	\sigma(\widetilde R_\ell^T W^{(\ell)}_* \widetilde R_\ell R_\ell x) 
\eeqn 
allows for a misalignment of the coordinate system \eqref{eq-layerell-basis-1-0} with the eigenbasis of $|W^{(\ell)}|$. This introduces additional degrees of freedom that account for a rotation, via $\widetilde R_\ell$, of the coordinate system in which $\sigma$ is defined.

Therefore, we introduce the following definition.

\begin{definition}\label{def-Wsig-aligned-1-0}
	We say that the cumulative weight matrix $W^{(\ell)}:\layersp_0\rightarrow\layersp_\ell$ is {\em aligned} with the activation function $\sigma$ if $|W^{(\ell)}|$ is diagonal in the coordinate system \eqref{eq-layerell-basis-1-0}, for $\ell=1,\dots,Q$. That is, 
\eqn 
	W^{(\ell)} = W^{(\ell)}_*  R_\ell
\eeqn 	
with $W^{(\ell)}_*$ diagonal, and $R_\ell\in O(Q)$. 
\end{definition}

We will see that for weight matrices aligned with the activation function, the gradient flow generated by the Euclidean loss in the input layer has a transparent form amenable to a clear understanding of the geometry of the minimization process via the dynamical reduction of the complexity of data clusters.


 
\section{Gradient flow in input space for cluster separated truncations}

In this section, we prove that gradient descent flow generated by the Euclidean loss is equivalent to a time dependent flow of truncation maps in input space $\layersp_0$ determined by the averages of input data clusters as they are progressively truncated.

\subsection{Definitions and notations}

We define the matrices
\eqn 
	X^{(\ell)}_j := [x_{j,1}^{(\ell)}\cdots x_{j,N_j}^{(\ell)}]
	\;\;\;,\;\;\;
	\Delta X^{(\ell)}_j := [\Delta x_{j,1}^{(\ell)}\cdots \Delta x_{j,N_j}^{(\ell)}]
\eeqn 
associated to the $j$-th class of data (associated to the reference output $y_j$), and
\eqn 
	X^{(\ell)} := [X_1^{(\ell)}\cdots X_Q^{(\ell)}] 
	\;\;\;,\;\;\;
	\Delta X^{(\ell)} := [\Delta X_1^{(\ell)}\cdots \Delta X_Q^{(\ell)}] \,.
\eeqn 
To begin with, we address the following special configuration of truncation maps and training data sets.

\begin{definition}
The presence of {\em cluster separated truncations} refers to the situation in which) the $\ell$-th truncation map $\tau^{(\ell)}$ acts as the identity on all clusters $\ell'\neq\ell$, for all $\ell=1,\dots,Q$. 
\end{definition}

Cluster separated truncations allow for zero loss optimization of the loss if the data are sufficiently clustered, see \cite{cheewa-2,cheewa-4}. We will discuss the geometry of the corresponding gradient flow in input space in detail; this will serve as the reference system for more general configurations in which truncation maps act on multiple clusters.

Given cluster separated truncations, we have that
\eqn\label{eq-tauinv-1-0}
	\tau^{(\ell)}(X_{\ell'}^{(0)}) = X_{\ell'}^{(0)}
	\;\;\;\forall \ell'\neq \ell \,.
\eeqn 
Then,
\eqn 
	\utau^{(Q)}(X^{(0)})&=& 
	\tau^{(Q)}\circ\cdots\circ\tau^{(1)}[X^{(0)}]
	\nonumber\\
	&=&
	\tau^{(Q)}\circ\cdots\circ\tau^{(2)}\big[
	[\tau^{(1)}(X^{(0)}_1)\dots \tau^{(1)}(X^{(0)}_\ell)\dots
	\tau^{(1)}(X^{(0)}_Q)] 
	\big]
	\nonumber\\
	&=&
	\tau^{(Q)}\circ\cdots\circ\tau^{(2)}\big(
	[\tau^{(1)}[X^{(0)}_1]\dots X^{(0)}_\ell\dots
	X^{(0)}_Q] 
	\big)
	\nonumber\\
	&=&\cdots\;=\;
	[\tau^{(1)}(X^{(0)}_1)\dots \tau^{(\ell)}(X^{(0)}_\ell)\dots
	\tau^{(Q)}(X^{(0)}_Q)] \,,
\eeqn 
that is, in particular,
\eqn 
	\utau^{(Q)}(X_\ell^{(0)}) = \tau^{(\ell)}(X_\ell^{(0)}) \,,
\eeqn 
and the Euclidean loss yields
\eqn 
	\CostN &=&\frac12 \sum_{\ell=1}^Q \frac1{N_\ell}\sum_{i=1}^{N_\ell} \Big|
	\tau^{(\ell)}(x^{(0)}_{\ell,i})-(W^{(Q+1)})^{-1}y_\ell  
	\Big|^2
	\\ 
	&=&\frac12 \sum_{\ell=1}^Q \frac1{N_\ell} \tr\Big(\Big|
	\tau^{(\ell)}(X^{(0)}_\ell)-(W^{(Q+1)})^{-1}y_\ell u_{N_\ell}^T
	\Big|^2\Big)
	\\
	&=&\frac12 \sum_{\ell=1}^Q \frac1{N_\ell} \tr\Big(\Big|
	\Delta\tau^{(\ell)}(X^{(0)}_\ell)
	\Big|^2\Big)
	+ \frac12\sum_{j=1}^Q 
	\Big|\overline{\tau^{(\ell)}(\{x^{(0)}_{\ell,i}\})} -(W^{(Q+1)})^{-1}y_\ell\Big|^2
	\nonumber
\eeqn 
where $u_{N_\ell}:=(1,1,\dots,1,1)^T\in\R^{N_\ell}$, and
\eqn 
	\overline{\tau^{(\ell)}(\{x^{(0)}_{\ell,i}\})}
	:=\frac1{N_\ell}\sum_{i=1}^{N_\ell} \tau^{(\ell)}(x^{(0)}_{\ell,i})
\eeqn 
denotes the $\ell$-th cluster average of the truncated data.

\subsection{Gradient flow for Euclidean cost}
We will first determine the gradient flow with respect to the {\em cumulative parameters} \eqref{eq-cumulWB-def-1-0} of the standard loss $\widetilde\CostN$ in \eqref{eq-CostNpullb-def-1-0}. We observe that  the truncation map $\tau^{(\ell)}$ depends on $(W^{(\ell')},\beta^{(\ell')})$ only for $\ell'=\ell$.

We introduce the following notations, for $x\in\R^Q$ and associated to $\sigma(x)=x_+$ denoting the ReLU activation,
\eqn 
	H(x) &:=& \diag(h(x_1),\dots,h(x_Q))
	\;\;\;,\;\;\;
	h(x_i)= \left\{
	\begin{array}{cc}
		1 & x_i>0 \\
		0 & x_i\leq 0
	\end{array} \right. \,,
	\nonumber\\
	H^\perp(x)&:=&\1_{Q\times Q}- H(x) \,,
\eeqn 
where $h$ is the Heaviside function.
Moreover, we let
\eqn 
	H_{W,\beta}(x) &:=& H(W(x+\beta)) \,,
	\nonumber\\
	H_{W,\beta}^\perp(x)&:=&\1_{Q\times Q}- H_{W,\beta}(x) \,
\eeqn
and 
\eqn
	H^{(\ell)}(x)&:=&H_{W^{(\ell)},\beta^{(\ell)}}(x)
	\nonumber\\
	H^{(\ell)\perp}(x)&:=&H_{W^{(\ell)},\beta^{(\ell)}}^\perp(x) \,,
\eeqn 
Then, using $H(x)x=\sigma(x)$, we obtain 
\eqn\label{eq-tau-H-id-1-0}
	\tau^{(\ell)}(x) = (W^{(\ell)})^{-1} H^{(\ell)}(x) W^{(\ell)} x 
	- (W^{(\ell)})^{-1} H^{(\ell) \perp}(x) W^{(\ell)}  \beta^{(\ell)}
\eeqn 
and
\eqn\label{eq-betaell-ODE-1-1-0}
	\partial_{\beta^{(\ell)}} \tau^{(\ell)}(x)
	&=& (W^{(\ell)})^{-1} \partial_{\beta^{(\ell)}}
	\sigma(W^{(\ell)}(x+\beta^{(\ell)})) - \partial_{\beta^{(\ell)}}\beta^{(\ell)}
	\nonumber\\
	&=& (W^{(\ell)})^{-1} H^{(\ell)}(x) W^{(\ell)} - \1_{Q\times Q} 
	\nonumber\\
	&=&-(W^{(\ell)})^{-1} H^{\perp (\ell)}(x) W^{(\ell)} 
\eeqn 
for $x\in\R^Q$.

We denote the empirical probability distribution on $\R^Q$, associated to the $\ell$-th cluster of training inputs, by
\eqn\label{eq-emp-probmeas-1-0}
	\mu_\ell(x):= \frac1{N_\ell}\sum_{i=1}^{N_\ell} 
	\delta(x-x_{\ell,i}^{(0)}) \,,
\eeqn
where $\delta$ is the Dirac delta distribution. Then, 
\eqn 
	\CostN &=&\frac12 \sum_{\ell=1}^Q 
	\int_{\R^Q} dx \, \mu_{\ell}(x)
	\, \Big|
	\tau^{(\ell)}(x)-(W^{(Q+1)})^{-1}y_\ell  
	\Big|^2 \,.
\eeqn
We then obtain the explicit gradient flow generated by the Euclidean loss in the input space $\layersp_0$ in the following theorem.  

\begin{theorem}\label{thm-gradflow-input-1-0}
	Let $\ell\in\{1,\dots,Q\}$, and
\eqn
	\widetilde y_\ell := (W^{(Q+1)})^{-1}y_\ell 
\eeqn
for notational convenience, with $W^{(Q+1)}$ fixed. 

Assume that that the $\ell$-th truncation map $\tau^{(\ell)}$ acts as the identity on all clusters $\ell'\neq\ell$ so that \eqref{eq-tauinv-1-0} holds, and that $W^{(\ell)}$ and $\sigma$ are aligned. Then, we may assume without any loss of generality that
\eqn 
	W^{(\ell)} = R_\ell \in O(Q)\,.
\eeqn 
Let 
\eqn
	a_{R_\ell,\beta^{(\ell)}}(x):=R_\ell(x+\beta^{(\ell)}) 
	\;\;\;,\;\;\;
	a_{R_\ell,\beta^{(\ell)}}^{-1}(x)=R_\ell^T x-\beta^{(\ell)} \,
\eeqn
denote the affine map associated to the $\ell$-th hidden layer and its inverse, as in \eqref{eq-affine-def-1-0-0}.
It follows that the gradient flow for the cumulative biases is determined by
\eqn\label{eq-bell-ODE-1-1-0} 
	\partial_s\beta^{(\ell)}
	&=&-\partial_{\beta^{(\ell)}} \CostN 
	\nonumber\\
	&=&  
	-R_\ell^T \Big(\int_{\R^Q\setminus\R_+^Q}dx \, 
	\mu_{\ell}(a_{R_\ell,\beta^{(\ell)}}^{-1}(x)) 
	  H^{\perp}(x)\Big)
	 R_\ell  
	( \beta^{(\ell)}+\widetilde y_\ell) \,.
\eeqn 
Moreover, let
\eqn
	\pi_- : \R^{Q\times Q} \rightarrow o(Q)
	\;\;\;,\;\;\;
	A \mapsto \frac12(A-A^T)
\eeqn 
denote the projection of $\R^{Q\times Q}$ to the Lie algebra $o(Q)$ of $O(Q)$  (i.e., the $\R$-linear subspace of antisymmetric matrices). Then, the gradient flow for the cumulative weights $R_\ell$, $\ell=1,\dots,Q$, is determined by  
\eqn\label{eq-Rell-ODE-1-1-0}
	\partial_s R_\ell(s) 
	=\,\Omega_\ell(s)  R_\ell(s)
\eeqn
with
\eqn\label{eq-Om-commut-def-1-0}
	\Omega_\ell
	&:=&-\pi_-((\partial_{R_\ell} \CostN)R_\ell^T )
	\nonumber\\
	&=&
	\int_{\R^Q\setminus(\R_+^Q\cup \R_-^Q)}  
	dx \, \mu_\ell(a_{R_\ell,\beta^{(\ell)}}^{-1}(x))
	\big[H(x)\;,\;
	M^{(\ell)}(x) \big] \,,
\eeqn
where $[A,B]=AB-BA$ is the commutator of $A,B\in\R^{Q\times Q}$, and
\eqn
	M^{(\ell)}(x) :=  
	\frac12\Big(x (\beta^{(\ell)}+\widetilde y_\ell)^T R_\ell^T 
	+ R_\ell(\beta^{(\ell)}+\widetilde y_\ell) x^T\Big)
	\,.
\eeqn
Along orbits of the gradient flow, the loss is monotone decreasing
\eqn\label{eq-CostN-monotone-1-0-0}
	\partial_s\CostN = 
	\sum_{\ell=1}^Q  
	\Big(
	(\partial_{\beta^{(\ell)}}\CostN)\cdot \partial_s\beta^{(\ell)}
	+ \tr\Big((\partial_{R_\ell}\CostN)^T\partial_s R_\ell\Big)
	\;\Big) \;\leq\;0\,,
\eeqn 
where in particular, both
\eqn
	(\partial_{\beta^{(\ell)}}\CostN)\cdot \partial_s\beta^{(\ell)}
	&=& - \Big|R_\ell^T 
	\Big(
	\int
	dx \mu_\ell(a_{R_\ell,\beta^{(\ell)}}^{-1}(x))
	  H^{\perp}(x)\Big)
	 R_\ell  
	( \beta^{(\ell)}+\widetilde y_\ell)\Big|^2
	\nonumber\\
	&\leq& 0
\eeqn 
and
\eqn\label{eq-derRCostN-monotone-1-0-0}
	\tr\Big((\partial_{R_\ell}\CostN)^T\partial_s R_\ell\Big)
	= - \tr\big(\;|\Omega_\ell|^2\;\big)
	\;\leq\;0
\eeqn 
are separately negative semidefinite for every $\ell=1,\dots,Q$.
\end{theorem}

The proof is given in Section \ref{sec-prf-thm-gradflow-input-1-0}.

We remark that for any symmetric matrix $M=M^T\in\R^{Q\times Q}$, 
\eqn
	\lefteqn{
	H(x)M-MH(x)
	}
	\nonumber\\
	&=&H(x)M(H(x)+H^\perp(x))-(H(x)+H^\perp(x))M H(x)
	\nonumber\\
	&=&H(x) M H^\perp(x)- H^\perp(x) M H(x)
\eeqn
by diagonality (and hence symmetry) of $H(x)$.

\subsection{Gradient flow and moments of $\mu_\ell$}
We make the key observation that the distribution of training inputs determines the gradient flow only via the zeroth and first free and constrained moments of $\mu_\ell$.

\begin{definition}
Given $\unu\in\{0,1\}^Q$, we define the $\unu$-th sector 
\eqn 
	\R_{\unu}^Q:=\{x\in\R^Q \;|\; h(x_i)=\nu_i\}
\eeqn 	
where the statement $h(x_i)=\nu_i$ is equivalent to $x_i>0$ if $\nu_i=1$, and $x_i\leq 0$ if $\nu_i=0$, for $i=1,\dots,Q$. 
\end{definition}

\begin{definition}[Free and constrained moments of $\mu_\ell$]
	Let $\mu_\ell$ denote the probability distribution \eqref{eq-emp-probmeas-1-0} associated to training data $\{x_{\ell,i}^{(0)}\}$. We define the free moments of zeroth and first degrees,
\eqn
	I_0^{(\ell)}&:=&\int_{\R^Q}dx \, 
	\mu_{\ell}(a_{R_\ell,\beta^{(\ell)}}^{-1}(x)) \; 
	\nonumber\\
	I_1^{(\ell)}&:=&\int_{\R^Q}dx \, 
	\mu_{\ell}(a_{R_\ell,\beta^{(\ell)}}^{-1}(x)) \; x
\eeqn
the constrained moments of zeroth degree,
\eqn
	J_0^{(\ell)}&:=&\int_{\R^Q}dx \, 
	\mu_{\ell}(a_{R_\ell,\beta^{(\ell)}}^{-1}(x)) \; 
	H(x)
\eeqn
and the first degree moments constrained to the $\unu$-th sector,
\eqn 
	\nonumber\\
	J_{1, \unu}^{(\ell)}&:=&\int_{\R^Q_\unu}dx \, 
	\mu_{\ell}(a_{R_\ell,\beta^{(\ell)}}^{-1}(x)) \;  
	  \; x \,.
\eeqn 
Moreover, 
\eqn
	J_0^{(\ell)\perp}&:=&I_0^{(\ell)}\1-J_0^{(\ell)}
	\; = \; \int_{\R^Q}dx \, 
	\mu_{\ell}(a_{R_\ell,\beta^{(\ell)}}^{-1}(x)) \; 
	H^\perp(x)
\eeqn
We note that $I_0^{(\ell)}$ is a scalar, while $J_0^{(\ell)}$ is a diagonal matrix.
\end{definition}

We note that the $r$-th diagonal component of $J_0^{(\ell)}$, $r=1,\dots,Q$,
\eqn
	(J_0^{(\ell)})_{rr}&=&\int_{\R^Q}dx \, 
	\mu_{\ell}(a_{R_\ell,\beta^{(\ell)}}^{-1}(x)) \; 
	h(x_r)
	\nonumber\\
	&=&\int_{\{x\in\R^Q|x_r>0\}}dx \, \mu_\ell(a_{R_\ell,\beta^{(\ell)}}^{-1}(x))
\eeqn 	
(where we recall that $h(x)$ is the Heaviside function) is the measure of the half space $\{x\in\R^Q|x_r>0\}$ with respect to the pullback probability density $\mu_\ell\circ a_{R_\ell,\beta^{(\ell)}}^{-1}$ under the affine map $a_{R_\ell,\beta^{(\ell)}}$.

We make the observation that
\eqn 
	H(x) = \diag(\unu) 
	\;\;\;,\;\;\;
	\forall x\in \R_{\unu}^Q \,,
\eeqn  
and that for any symmetric matrix $M=M^T\in\R^{Q\times Q}$,
\eqn 
	[H(x),M]_{ij} = (\nu_i-\nu_j)M_{ij}
	\;\;\;,\;\;\;
	\forall x\in \R_{\unu}^Q 
\eeqn 
for all $i,j\in\{1,\dots,Q\}$. 
Therefore, we can write the gradient flow equations for $\beta^{(\ell)}(s)$ and $R_\ell(s)$ in the following form.

\begin{corollary}
\label{cor-gradflow-1-0-0}
	We make the same assumptions as in Theorem \ref{thm-gradflow-input-1-0}. Expressed in terms of moments of $\mu_\ell$, the gradient flow equations \eqref{eq-betaell-ODE-1-1-0} and \eqref{eq-Rell-ODE-1-1-0} for the cumulative biases and weights $\beta^{(\ell)}$ and $W^{(\ell)}=W^{(\ell)}_*R_\ell$, with $W^{(\ell)}_*\geq0$ diagonal, are given by
\eqn\label{eq-beta-ODE-moments-1-0}
	\partial_s\beta^{(\ell)}
	=  
	-R_\ell^T J_0^{(\ell)\perp}
	 R_\ell  
	( \beta^{(\ell)}+\widetilde y_\ell) \,,
\eeqn
and 
\eqn 
	\partial_{s}R_\ell(s) = \Omega_\ell(s) R_\ell(s)
\eeqn 
where the matrix elements of $\Omega_\ell$ are given by
\eqn\label{eq-Omell-ij-mom-1-0-0}
	[\Omega_\ell]_{ij} 
	&=&  
	\sum_{\unu\in\{0,1\}^Q}\int_{\R_{\unu}^Q} dx
	\mu_\ell(a_{R_\ell,\beta^{(\ell)}}^{-1}(x))
	(\nu_i-\nu_j)
	\nonumber\\
	&&
	\hspace{1cm}
	\Big(x_i \, (R_\ell (\beta^{(\ell)}+\widetilde y_\ell))_j
	+ (R_\ell (\beta^{(\ell)}+\widetilde y_\ell))_i \, x_j \Big)
	\nonumber\\
	&=&\sum_{\unu\in\{0,1\}^Q} 
	(\nu_i-\nu_j)
	\\
	&&
	\hspace{1cm}
	\Big((J_{1, \unu}^{(\ell)})_i \, 
	(R_\ell (\beta^{(\ell)}+\widetilde y_\ell))_j
	+ (R_\ell (\beta^{(\ell)}+\widetilde y_\ell))_i \, 
	(J_{1, \unu}^{(\ell)} )_j \Big)
	\nonumber
\eeqn 
for $\ell=1,\dots,Q$ and $i,j\in\{1,\dots,Q\}$. In particular, the gradient flow depends on the training inputs only through the first degree moments of $\mu_\ell\circ a_{R_\ell,\beta^{(\ell)}}^{-1}$ restricted to the sectors $\R_{\unu}^Q$.
\end{corollary}

We note that the $r$-th component of $J_0^{(\ell) \perp}$ in \eqref{eq-beta-ODE-moments-1-0} is the measure of the complementary half space $\{x\in\R^Q|x_r\leq 0\}$ with respect to $\mu_\ell\circ a_{R_\ell,\beta^{(\ell)}}^{-1}$. Explicitly,
\eqn 
	J_0^{(\ell)\perp} = 
	\diag\Big(\frac{n^{(\ell)}_1}{N_\ell},\dots,
	\frac{n^{(\ell)}_Q}{N_\ell}\Big) \,,
\eeqn
where
\eqn\label{eq-def-nrell-1-0-0}
	n_r^{(\ell)} =\#\Big\{x_{\ell,i}^{(0)}\in\R^Q
	\;,\;i=1,\dots,N_\ell \;\Big|
	\; \Big(a_{R_\ell,\beta^{(\ell)}}(x_{\ell,i}^{(0)})\Big)_r\leq 0\Big\}
\eeqn 
is the number of training data $x_{\ell,i}^{(0)}$ for which the $r$-th component of $a_{R_\ell,\beta^{(\ell)}}(x_{\ell,i}^{(0)})$ is negative. Therefore, the gradient flow for $\beta^{(\ell)}(s)$ is driven by the number of training inputs $x_{\ell,i}^{(0)}$ that have been truncated by $\tau^{(\ell)}$ as time elapses.

Furthermore, we note that in agreement with \eqref{eq-Om-commut-def-1-0}, the contributions from $\R_+^Q$ (where $\nu_i=1$ for all $i$) and $\R_-^Q$ (where $\nu_i=0$ for all $i$) to \eqref{eq-Omell-ij-mom-1-0-0} are zero because $\nu_i-\nu_j=0$ for all $i,j$ in both cases.




To solve the system of ODEs \eqref{eq-bell-ODE-1-1-0} and \eqref{eq-Rell-ODE-1-1-0}, {\rm no backpropagation is needed}; that is, the Jacobi matrix of the map from parameter space to the output space does not need to be calculated for each time step. 

In Section \ref{sec-geom-illustr-1-0}, we will further elucidate the geometric interpretation of the flow of cumulative biases and weights.

\section{Geometry of orbits for cluster separated truncations}
\label{sec-geom-illustr-1-0}

In this section, we discuss the geometric interpretation of the gradient flow in input space, as presented in Theorem \ref{thm-gradflow-input-1-0}.

We use the expressions for the gradient flow equations as given in Corollary \ref{cor-gradflow-1-0-0},
\eqn\label{eq-gradflow-moments-1-0-1}
	\partial_s(\beta^{(\ell)}+\widetilde y_\ell)
	&=&  
	 - R_\ell^T J_0^{(\ell) \perp} R_\ell
	( \beta^{(\ell)}+\widetilde y_\ell) \,
	\nonumber\\ 
	\partial_s R_\ell 
	&=& - \, \Omega_\ell  R_\ell
\eeqn 
where
\eqn
	J_0^{(\ell) \perp} = \int_{
	\R^Q\setminus\R^Q_+}
	dx \, \mu_\ell(a_{R_\ell,\beta^{(\ell)}}^{-1}(x)) \, H^\perp(x) \,,
\eeqn
and
\eqn\label{eq-Omell-ij-mom-1-0-1}
	[\Omega_\ell]_{ij} &=& 
	\sum_{\unu\in\{0,1\}^Q}\int_{
	\R_{\unu}^Q\subset\R^Q\setminus(\R^Q_+\cup \R^Q_-)}
	dx
	\, \mu_\ell(a_{R_\ell,\beta^{(\ell)}}^{-1}(x))
	(\nu_i-\nu_j)
	\nonumber\\
	&&
	\hspace{1cm}
	\Big(x_i \, (R_\ell (\beta^{(\ell)}+\widetilde y_\ell))_j
	+ (R_\ell (\beta^{(\ell)}+\widetilde y_\ell))_i \, x_j \Big)
\eeqn 
We present several cases in which the gradient flow can be explicitly controlled.

\subsection{Equilibria}
We straightforwardly obtain equilibrium solutions in the following two cases, which coincide with the stationary solutions determined in \cite{cheewa-2,cheewa-4} using variational arguments.

\begin{proposition}
\label{prop-equi-0-1}
	Assume that the parameters $R_{\ell *}\in O(Q)$ and $\beta^{(\ell)}_*\in\R^Q$ are such that the training data $\{x_{\ell,i}^{(0)}\}_{i=1,\dots,N_\ell}$ are either fully untruncated,
	\eqn 
		\supp(\mu_\ell) \subset \cS_\ell^+
	\eeqn 
	or fully truncated,
	\eqn 
		\supp(\mu_\ell) \subset \cS_\ell^- \,
	\eeqn 
	(see \eqref{eq-cSell-def-1-0-0} for definitions).
	Then, in either case, $(\beta^{(\ell)}_*,R_{\ell *})$ is an equilibrium solution to the gradient flow.
\end{proposition}

\prf 

\noindent
{\em Case 1: Initial data fully untruncated.}
Assume at initial time that $R_\ell(0)$ and $\beta^{(\ell)}(0)$ are such that 
\eqn
	\supp(\mu_\ell) \subset \cS_\ell^+ \,.
\eeqn
This corresponds to $\tau^{(\ell)}(x_{\ell,i}^{(0)})=x_{\ell,i}^{(0)}$ for all $i=1,\dots,N_\ell$, which is equivalent to
\eqn
	\cM:=
	\supp\Big(\mu_\ell\circ a_{R_\ell(0),\beta^{(\ell)}(0)}^{-1}\Big) 
	\subset \R_+^Q \,.
\eeqn
This in turn is equivalent to $J_0^{(\ell) \perp}=0$, which implies $H^\perp(x)=0$ and $H(x)=\1$ for all $x\in\cM$. It then follows that $\Omega_\ell=0$ because in \eqref{eq-Om-commut-def-1-0}, $H(x)=\1$ trivially commutes with $M_\ell(x)$. This in turn implies that $\partial_s\beta^{(\ell)}(s)=0$ and $\partial_s R_\ell=0$.

\noindent
{\em Case 2: Initial data fully truncated.}
Assume at initial time that $R_\ell(0)$ and $\beta^{(\ell)}(0)$ are such that 
\eqn
	\supp(\mu_\ell) \subset \cS_\ell^- \,.
\eeqn 
This corresponds to $\tau^{(\ell)}(x_{\ell,i}^{(0)})=-\beta^{(\ell)}$ for all $i=1,\dots,N_\ell$, which is equivalent to
\eqn
	\cM=
	\supp\Big(\mu_\ell\circ a_{R_\ell(0),\beta^{(\ell)}(0)}^{-1}\Big) 
	\subset \R_-^Q \,.
\eeqn
Then, $J_0^{(\ell) \perp}=I_0^{(\ell)}$, and $H^\perp(x)=\1$ and $H(x)=0$ for all $x\in\cM$. It then follows that $\Omega_\ell=0$ because in \eqref{eq-Om-commut-def-1-0}, $H(x)=0$. This implies that $\partial_s\beta^{(\ell)}(s)=0$ and $\partial_s R_\ell=0$.
\endprf

\subsection{Flow of $\beta^{(\ell)}$ and $R_\ell$ for partially truncated initial data}

Assume at initial time that $R_\ell(0)$ and $\beta^{(\ell)}(0)$ are such that
\eqn
	\cM=
	\supp\Big(\mu_\ell\circ 
	a_{R_\ell(0),\beta^{(\ell)}(0)}^{-1}\Big) 
	\cap (\R^Q\setminus \R_-^Q) \neq \emptyset \,,
\eeqn
so that $(J_0^{(\ell) \perp})_r>0$ for each component $r=1,\dots,Q$; that is, the cluster of training data is partially truncated in all coordinate directions. 


Moreover, we observe that the integration domain
$\R^Q\setminus\R^Q_+$  
of $J_0^{(\ell) \perp}$ that determines the flow of $\beta^{(\ell)}(s)$ contains the negative sector $\R^Q_-$, while the integration domain $\R^Q\setminus(\R^Q_+\cup\R^Q_-)$ 
of $\Omega_\ell$ consists only of the "off-diagonal" sectors excluding $\R^Q_+$ and $\R^Q_-$. This is important because we will see that as time elapses, the lower bound on $J_0^{(\ell) \perp}$ increases by moving the support of $\mu_\ell\circ a_{R_\ell,\beta^{(\ell)}}^{-1}$ into the negative sector $\R^Q_-$, while $\Omega_\ell$ converges to zero. Thereby, $\beta^{(\ell)}(s)$ converges to $-\widetilde y_\ell$ exponentially at a rate that increases as time elapses, while $R_\ell(s)$ becomes stationary.

As regards the latter, is instructive to address the a priori asymptotics of $R_\ell\in O(Q)$ in the limit $s\rightarrow\infty$. The loss converges in the limit $s\rightarrow\infty$, due to monotone decrease along orbits \eqref{eq-CostN-monotone-1-0-0}, and boundedness below, $\CostN\geq0$. Therefore, $\partial_s\CostN\leq0$ as $s\rightarrow\infty$, along orbits of the gradient flow.
From \eqref{eq-derRCostN-monotone-1-0-0},  
\eqn 
	\partial_s\CostN \leq -\tr(|\Omega_\ell(s)|^2) \leq0 \,,
\eeqn 
and $\partial_s\CostN\rightarrow0$ implies that in operator norm, $\|\Omega_\ell\|\rightarrow0$ as $s\rightarrow \infty$, which in turn implies that
\eqn
	\|\partial_sR_\ell(s)\|
	\leq \|\Omega_\ell(s)\|
	\, \|R_\ell\|\rightarrow 0
\eeqn 
(by orthogonality, $\|R_\ell\|=1$) as $s\rightarrow\infty$. 

In Theorem \ref{eq-gradflow-conv-1-0-0}, we make the above discussion rigorous.



\begin{theorem}
\label{eq-gradflow-conv-1-0-0}
Let $0<\eta_0,\eta_1,\gamma<\frac1{10}$ be small constants, and 
\eqn
	\eta_1'&:=& 1.1  |\beta^{(\ell)}(0)+\widetilde y_\ell| 
	\;\eta_1
	\nonumber\\
	\eta_2 &:=&
	\frac{\eta_1}{1-\eta_0-\eta_1'}\log\frac1\gamma \,.
\eeqn
Assume that the initial data $(\beta^{(\ell)}(0),R_\ell(0))\in\R^Q\times O(Q)$ and the probability distribution 
$\mu_\ell
$ 
satisfy: 
\begin{itemize}
\item
mass concentration in the complement of the positive sector,
\eqn\label{eq-massconc-ass-1-0-0}
	\int_{\R^Q\setminus\R_+^Q}\mu_\ell\circ a_{R,\beta}(x) 
	H^\perp(x)
	> 1-\eta_0
\eeqn
for all $(\beta,R)\in\R^Q\times O(Q)$ satisfying  $|\beta+\widetilde y_\ell|\leq 1.1|\beta^{(\ell)}(0)+\widetilde y_\ell|$ and $\|R-R_\ell(0)\|<\eta_2$.
\item
small first degree moment in off-diagonal sectors
\eqn
	\int_{\R^Q\setminus(\R_+^Q\cup\R_-^Q)}
	\mu_\ell\circ a_{R,\beta}(x) |x|
	< \eta_1
\eeqn
for all $(\beta,R)\in\R^Q\times O(Q)$ satisfying  $|\beta+\widetilde y_\ell|\leq 1.1|\beta^{(\ell)}(0)+\widetilde y_\ell|$ and $\|R-R_\ell(0)\|<\eta_2$.
\item
translation of the entire mass into the negative sector for $\beta$ close enough to $\widetilde y_\ell$; that is,
\eqn
	\supp\Big(\mu_\ell\circ 
	a_{R,\beta}^{-1}\Big)\subset\R_-^Q
\eeqn
for all $\beta\in\R^Q$ satisfying
\eqn
	|\beta+\widetilde y_\ell| <
	\gamma |\beta^{(\ell)}(0)+\widetilde y_\ell| \,.
\eeqn
and all $R\in O(Q)$ with $\|R-R_\ell(0)\|<\eta_2$.
\end{itemize}
Then, the solution to the gradient flow \eqref{eq-gradflow-moments-1-0-1} with initial data $(\beta^{(\ell)}(0),R_\ell(0))$ satisfies the following. 

There exists a finite time $0<s_1<\infty$ such 
\eqn
	\supp\Big(\mu_\ell\circ 
	a_{R_\ell(s_1),\beta^{(\ell)}(s_1)}^{-1}\Big)\subset\R_-^Q
\eeqn
and
\eqn
	\Omega_\ell(s_1) = 0 \,.
\eeqn
For $s>s_1$,
\eqn
	|\beta^{(\ell)}(s)+\widetilde y_\ell| < e^{-(s-s_1)}
	|\beta^{(\ell)}(s_1)+\widetilde y_\ell|
\eeqn
converges to zero as $s\rightarrow\infty$, and
\eqn 
	R_\ell(s)= R_\ell(s_1)
\eeqn
holds.

\end{theorem}

\prf
With
\eqn
	\tb^{(\ell)} := R_\ell (\beta^{(\ell)}+\widetilde y_\ell) \,,
\eeqn
we obtain
\eqn 
	\partial_s \tb^{(\ell)} &=&
	\underbrace{((\partial_s R_\ell)R_\ell^T)}_{=\Omega_\ell} 
	\, \tb^{(\ell)}
	+R_\ell\partial_s \beta^{(\ell)}\,
	\nonumber\\
	&=&
	\Omega_\ell \, \tb^{(\ell)}
	- J_0^{(\ell) \perp} \, \tb^{(\ell)}
\eeqn 
By assumption \eqref{eq-massconc-ass-1-0-0}, we have that as long as $\beta^{(\ell)}(s)$ satisfies $|\beta^{(\ell)}(s)+\widetilde y_\ell|<1.1|\beta^{(\ell)}(0)+\widetilde y_\ell|$,
\eqn
	J_0^{(\ell) \perp} > 1-\eta_0
\eeqn
and recalling \eqref{eq-Omell-ij-mom-1-0-1}, we find that in operator norm,
\eqn 
	\|\Omega_\ell(s)\| &\leq& 
	\Big(\int_{\R^Q\setminus(\R^Q_+\cup\R^Q_-)}
	dx \mu_\ell(a_{R_\ell(s),\beta^{(\ell)}(s)}^{-1}(x)) |x|
	\Big)
	|\tb^{(\ell)}|
	\nonumber\\
	&<&
	\eta_1 |\tb^{(\ell)}|
	\nonumber\\
	&<&
	1.1  |\tb^{(\ell)}(0)| \; \eta_1
	\;=:\;\eta_1' \,.
\eeqn 
Therefore,
\eqn 
	\|\partial_s R_\ell(s)\| =
	\|(\partial_s R_\ell(s))R_\ell^T(s)\| 
	\leq \|\Omega_\ell(s)\| \leq  \eta_1'\,.
\eeqn 
This implies that for all $(\tb^{(\ell)},R_\ell)\in\cU_{\eta_0,\eta_1}(\tb^{(\ell)}(0),R_\ell(0))$,
\eqn 
	J_0^{(\ell) \perp} - \Omega_\ell > 
	1 - \eta_0 - \eta_1'
\eeqn
and therefore,
\eqn 
	|\tb^{(\ell)}(s)| &<& e^{-s(1-\eta_0-\eta_1')}
	|\tb^{(\ell)}(0)|
\eeqn 
and
\eqn 
	\|R_\ell(s)-R_\ell(0)\| < \eta_1' s
\eeqn 
which implies that there exists a finite time 
\eqn
	0< s_1 \leq   \frac{1}{1-\eta_0-\eta_1'}\log\frac1\gamma
\eeqn
such that 
\eqn 
	|\tb^{(\ell)}(s_1)| < 
	\gamma |\beta^{(\ell)}(0)+\widetilde y_\ell|
\eeqn 
and
\eqn 
	\|R_\ell(s_1)-R_\ell(0)\| &<& \eta_1' s_1 
	\nonumber\\
	&<& \eta_2 :=
	\frac{\eta_1'}{1-\eta_0-\eta_1}\log\frac1\gamma
\eeqn
where
\eqn 
	\supp \Big(\mu_\ell(a_{R_\ell(s_1),\beta^{(\ell)}(s_1)})\Big)
	\subset\R^Q_- \,.
\eeqn 
Therefore, we have that
\eqn 
	\Omega_\ell(s_1) = 0  
\eeqn 
and
\eqn
	J_0^{(\ell) \perp}(s_1)=\1 \,.
\eeqn 
This implies that for $s>s_1$, the gradient flow equations reduce to 
\eqn 
	\partial_s \tb^{(\ell)} &=& 
	-  \tb^{(\ell)}
	\nonumber\\
	\partial_s R_\ell &=&0
\eeqn 
which implies that
\eqn 
	\tb^{(\ell)}(s) &=& e^{-(s-s_1)}\tb^{(\ell)}(s_1)
	\nonumber\\
	R_\ell(s)&=&R_\ell(s_1) \,.
\eeqn
Thus, asymptotically, $\tb^{(\ell)}(s)\rightarrow0$ as $s\rightarrow\infty$, or equivalently, $\beta^{(\ell)}(s)\rightarrow -\widetilde y_\ell$.

In particular, $|\tb^{(\ell)}(s)|<|\tb^{(\ell)}(0)|$ holds for all $s>0$, and therefore, the assumption $|\tb^{(\ell)}(s)|<1.1|\tb^{(\ell)}(0)|$ is satisfied for all $s\geq0$ along the orbit.
\endprf

In the next Section \ref{ssec-beta-ex-flow-1-0}, we present a detailed calculation which further elucidates the precise manner in which $\beta^{(\ell)}(s)$ converges to $-\widetilde y_\ell$ at an exponential rate that increases with the number of training data that are progressively truncated as time elapses.

\subsection{Flow of $\beta^{(\ell)}$ at fixed $R_\ell$}
\label{ssec-beta-ex-flow-1-0}
To understand in detail some key properties of the dynamics of orbits of the gradient flow, let us fix $R_\ell\in O(Q)$ to be constant, and only focus on the flow of the cumulative biases. This is motivated by the asymptotics of $R_\ell(s)$ just discussed as $s\rightarrow\infty$, if we assume $R_\ell$ to be convergent, and near a limiting value. Given $\ell\in\{1,\dots,Q\}$, and recalling the definition of the empirical probability density $\mu_\ell$ in \eqref{eq-emp-probmeas-1-0}, we have 
\eqn 
	\partial_s(\beta^{(\ell)}+\widetilde y_\ell)
	&=&-\partial_{\beta^{(\ell)}} \CostN 
	\nonumber\\
	&=&  
	-R_\ell^T \Big(\frac1{N_\ell} \sum_{i=1}^{N_\ell} 
	  H^{(\ell) \perp}(x_{\ell,i}^{(0)})\Big)
	 R_\ell  
	( \beta^{(\ell)}+\widetilde y_\ell) \,.
\eeqn 
Using
\eqn 
	b^{(\ell)}:=R_\ell \beta^{(\ell)} \,,
\eeqn 
we have
\eqn\label{eq-derb-def-1-0}
	\partial_s(b^{(\ell)}+R_\ell\widetilde y_\ell)
	&=&  
	- \Big(\frac1{N_\ell} \sum_{i=1}^{N_\ell} 
	  H^{(\ell) \perp}(x_{\ell,i}^{(0)})\Big)
	( b^{(\ell)}+R_\ell\widetilde y_\ell) \,,
\eeqn 
where we recall that 
\eqn 
	H^{(\ell) \perp}(x_{\ell,i}^{(0)}) = H^\perp(R_\ell x_{\ell,i}^{(0)} + b^{(\ell)})
\eeqn 
is diagonal. Hence, its $r$-th component depends only on the $r$-th component of $b^{(\ell)}$. The components of this ODE therefore decouple, and we may, without any loss of generality, restrict our analysis to one single component.  

For each $r\in\{1,\dots,Q\}$, the $r$-th component in the ODE \eqref{eq-derb-def-1-0} is a 1-dimensional problem of the form
\eqn
	\partial_s(y-b) = - \Big(\frac1{N_\ell} \sum_{i=1}^{N_\ell} 
	  h^{\perp}(x_i-b)\Big)
	( y-b)
\eeqn 
where we temporarily use the abbreviated notation
\eqn
	b:= -b^{(\ell)}_r
	\;\;\;,\;\;\;
	x_i := (R_\ell x_{\ell,i}^{(0)})_r
	\;\;\;,\;\;\;
	y:=(R_\ell\widetilde y_\ell)_r \,.
\eeqn 
Without any loss of generality, we assume that the data points $\{x_i\}_{i=1}^{N_\ell}\subset\R$ are labeled such that  $x_1< x_2 <\cdots<x_{N_\ell}$.
We recall that $h^{\perp}(x_i-b)=0$ if $x_i>b$, and $h^{\perp}(x_i-b)=1$ if $x_i\leq b$.
Therefore, if at initial time $s=0$, we have $x_i>b$ for all $i=1,\dots,N_\ell$, then the r.h.s. is zero, and we obtain a fixed point solution,
\eqn
	\partial_s(y-b) = 0 \,.
\eeqn 
Next, assume recursively that for $1\leq n\leq N_\ell$, the time $s=s_n$ is characterized by $x_1<x_2<\cdots<x_n = b(s_n)$. This means that  at time $s_n$, the training points $x_1,\dots,x_n$ have been truncated by $\tau^{(\ell)}$ in the $r$-th coordinate direction. Then, 
\eqn
	\frac1{N_\ell} \sum_{i=1}^{N_\ell} 
	  h^{\perp}(x_i-b) = \frac n{N_\ell} \,,
\eeqn 
and thus,
\eqn
	\partial_s(y-b) = -  \frac{n}{N_\ell} ( y-b)
\eeqn 
as long as $x_{n+1}>b$.	Clearly, 
\eqn\label{eq-yb-convexp-1-0}
	(y-b)(s)=e^{-\frac n {N_\ell}(s-s_n)}(y-b)(s_n)
\eeqn 
for $s\in [s_n,s_{n+1}]$, and hence, 
\eqn
	(y-b)(s_{n+1})&=&y-x_{n+1}
	\nonumber\\
	&=&
	e^{-\frac n {N_\ell}(s_{n+1}-s_n)}(y-b)(s_n)
	\nonumber\\
	&=& 
	e^{-\frac n {N_\ell}(s_{n+1}-s_n)}(y-x_n) \,.
\eeqn 
This implies that
\eqn 
	s_{n+1}=s_n+ \frac {N_\ell}n \log\frac{y-x_{n}}{y-x_{n+1}} \,.
\eeqn 
That is, once the $n$-th training point has been truncated, the $n+1$-st training point will be reached in finite time, for every $n\geq 1$. As $n$ increases, the exponential rate in \eqref{eq-yb-convexp-1-0} increases. Moreover, once all training points have been truncated, we find that
\eqn 
	(y-b)(s)&=&e^{-(s-s_{N_\ell})}(y-b)(s_{N_\ell})
	\nonumber\\
	&=&e^{-(s-s_{N_\ell})}(y-x_{N_\ell})
\eeqn 
for $s>s_{N_\ell}$. That is, $b(s)$ converges to $y$ exponentially, at the rate $1=\frac {N_\ell}{N_\ell}$. Here, $y$ corresponds to the pullback of the reference output vector to the input space.

\section{General gradient flow without cluster separated truncations}

In this section, we derive the explicit gradient flow equations for the cumulative weights and biases in which any truncation map may act nontrivially on any cluster, under the assumption that the weights are adapted to the activations. 

\begin{theorem}
\label{thm-gradflow-general-1-0}
Let for $\ell_2\geq\ell_1$,
\eqn 
	P^+_{\ell_1,\ell_2}(x)&:=& R_{\ell_2}^T H^{(\ell_2)} R_{\ell_2}
	\cdots
	R_{\ell_1}^T H^{(\ell_1)} R_{\ell_1}\,,
\eeqn  
and
\eqn 
	P^-_{\ell_1,\ell_2}(x) &:=& 
	R_{\ell_2}^T H^{(\ell_2)} R_{\ell_2} \cdots
	R_{\ell_1+1}^T H^{(\ell_1+1)} R_{\ell_1+1}
	R_{\ell_1}^T H^{(\ell_1) \perp } R_{\ell_1}
	\nonumber\\
	P^-_{\ell,\ell}(x) &:=& 
	R_{\ell}^T H^{(\ell) \perp} R_{\ell} \,,
\eeqn 
using  
\eqn 
	H^{(\ell)} \equiv H^{(\ell)}(\tau^{(\ell-1,1)}(x))
	=H(R_\ell(\tau^{(\ell-1,1)}(x)+\beta^{(\ell)} ) )
\eeqn
for notational brevity.

Then,
\eqn\label{eq-derb-full-1-0-0}
	\partial_s\beta^{(\ell)}
	&=&-\partial_{\beta^{(\ell)}}\CostN 
	\nonumber\\
	&=&-\sum_{\ell'=1}^Q
	\int
	dx \, \mu_{\ell'}(x)
	\frac12
	\partial_{\beta^{(\ell)}}
	\Big| \, \tau^{(Q,1)}(x)-\widetilde y \, \Big|^2
	\nonumber\\
	&=&
	-\sum_{\ell'=\ell}^Q
	\int
	dx \, \mu_{\ell'}(x)
	(P^-_{\ell',Q}(x))^T
	\Big(P^+_{\ell',Q}(x)\tau^{(\ell'-1,1)}(x)
	\nonumber\\
	&&\hspace{1cm}
	- \sum_{\ell''=\ell'}^{Q}
	P^-_{\ell'',Q}(x) \beta^{(\ell'')}
	-\widetilde y\Big) \,
\eeqn
determines the gradient flow of the cumulative biases. Moreover, 
\eqn 
	\partial_s R_\ell(s) = \widetilde\Omega_\ell(s) \, R_\ell(s)
\eeqn 
with 
\eqn
	\widetilde\Omega_\ell &=& -\pi_-\Big( 
	(\partial_{R_\ell}\CostN)R_\ell^T\Big)
	\nonumber\\
	&=&
	-\sum_{\ell'=1}^Q\int dx\,\mu_{\ell'}(x)[H^{(\ell)} ,
	M_{\ell,\ell'}(x)] 
\eeqn 
and
\eqn
	M_{\ell,\ell'}(x) &:=& \frac12 R_{\ell}\Big(
	(P^+_{\ell+1,Q})^T(\tau^{(Q,1)}(x)-\widetilde y_{\ell'})
	(\tau^{(\ell-1,1)}(x)+\beta^{(\ell)})^T  
	\\
	&&\hspace{1cm}
	+  (\tau^{(\ell-1,1)}(x)+\beta^{(\ell)})
	((P^+_{\ell+1,Q})^T(\tau^{(Q,1)}(x)-\widetilde y_{\ell'}))^T \Big)
	R_\ell^T \,
	\nonumber
\eeqn 
determines the gradient flow of the cumulative weights.
\end{theorem}

We observe that the integration domain of the $\ell'$-th term in \eqref{eq-derb-full-1-0-0},
\eqn 
	 \{x\in\R^Q \,| \, P^-_{\ell',Q}(x)\neq 0 \} \,,
\eeqn 
is contained in the intersection set
\eqn 
	\cD_{\ell',Q}^- :=  (\tau^{(\ell'-1,1)})^{-1}(\cS^\perp_{\ell'})
	\cap
	(\tau^{(\ell',1)})^{-1}(\cS^+_{\ell'+1})
	\cap\cdots\cap
	(\tau^{(Q-1,1)})^{-1}(\cS^+_Q) \,
\eeqn
where
\eqn 
	H^{(\ell) \perp} \neq 0
	\;,\;H^{(\ell+1)} \neq 0 \; , \; \cdots \; , \;
	H^{(Q)} \neq 0 \,.
\eeqn 
This can be understood in the sense that the concatenation of truncation maps renormalizes the integration domain.

Moreover, we note that clustering of training data corresponds to the property of the probability densities $\{\mu_\ell\}$ having pairwise disjoint supports,
\eqn 
	\supp(\mu_\ell)\cap\supp(\mu_{\ell'}) = \emptyset
	\;\;\;,\;\;\;
	\forall \ell\neq\ell' \,.
\eeqn
If cluster separated truncations exist for a given set of $\{\mu_\ell\}$, then they satisfy the condition that
\eqn
	R_\ell^T H^{(\ell)}(x)R_\ell x' = x'
	\;\;\;,\;\;\;
	\forall x\in\supp(\mu_\ell)\;
	{\rm and}\;x'\in\supp(\mu_{\ell'})\,.
\eeqn
Therefore, in this case, one finds that for all $\ell'\geq\ell$,
\eqn 
	\lefteqn{
	(P^-_{\ell',Q}(x))^T P^+_{\ell',Q}(x)\tau^{(\ell'-1,1)}(x)
	}
	\nonumber\\
	& =& 
	(P^-_{\ell',\ell'}(x))^T 
	P^+_{\ell',\ell'}(x)\tau^{(\ell'-1,1)}(x)
	\nonumber\\
	&=&0
\eeqn 
in agreement with Theorem \ref{thm-gradflow-input-1-0}.

An analysis of the flow equations derived in Theorem \ref{thm-gradflow-general-1-0} will be addressed in future work.

\section{Gradient flow for standard loss and collapsed initial data}
\label{sec-std-gradflow-1-0-0}

In this section and the next, we discuss the gradient flow for the {\em standard loss} \eqref{eq-CostNpullb-def-1-0} defined via the pullback metric in two special situations. First, we address the case in which the initial data are neurally collapsed, for $\sigma$ being ReLU as before. The issue of degeneracy described in Section \ref{ssec-standcost-1-0-0} will emerge as an aspect of our analysis.

To this end, we consider clustered training data where $x_{j,i}^{(0)}\in B_\delta(\overline{x_j}^{(0)})$ for $j=1,\dots,Q$, and $i=1,\dots,N_j$, for some sufficiently small $\delta>0$. Then, we choose initial data $(W^{(\ell)}(s=0),\beta^{(\ell)}(s=0))_\ell$ in a manner that
\eqn 	
	\tau^{(Q)}\big|_{s=0}(x^{(0)}_{\ell,i}) = -\beta^{(\ell)}
	\;\;\;,\;
	i=1,\dots,N_\ell 
	\;\;\;,\;
	\ell=1,\dots,Q \,.
\eeqn 	
That is, the $\ell$-th cluster is mapped to the point $-\beta^{(\ell)}$, for every $\ell=1,\dots,Q$.
The explicit construction of initial data satisfying these properties is presented in \cite{cheewa-2,cheewa-4}.

Then, the time-dependent standard loss reduces to
\eqn 
	\widetilde\CostN &=& \frac12\sum_{\ell=1}^Q |-W_{Q+1}(s)\beta^{(\ell)}(s)-y_\ell|^2
	\nonumber\\
	&=&\frac12\tr\Big(\big|-W_{Q+1}(s)B^{(Q)}(s)-Y\big|^2\Big)
\eeqn 
where we have set $b_{Q+1}=0$, and   
\eqn
	B^{(Q)}(s)&:=&[\beta^{(1)}(s)\cdots\beta^{(Q)}(s)]
	\nonumber\\
	Y&:=&[y_1\cdots y_Q]\,,
\eeqn
both in $\R^{Q\times Q}$.
The loss is independent of $(W^{(\ell)}(s=0))_\ell$, and this fact persists for $s>0$, because the gradient flow is trivial for the cumulative weights,
\eqn 
	\partial_s W^{(\ell)}(s) &=& -\partial_{(W^{(\ell)}(s))^T}
	\widetilde\CostN 
	\nonumber\\
	&=& 0 
	\;\;\;,\;\ell=1,\dots,Q \,.
\eeqn 
Thus, we find for $\ell=1,\dots,Q$,
\eqn 
	\partial_s \beta^{(\ell)}(s) &=& - \partial_{\beta^{(\ell)}}
	\widetilde\CostN
	\nonumber\\
	&=& - (W_{Q+1}(s))^T (W_{Q+1}(s) \beta^{(\ell)}(s) + y_\ell) \,,
\eeqn 
respectively, in matrix form,
\eqn\label{eq-derB-GD-1}
	\partial_s B^{(Q)}(s) &=& - \partial_{(B^{(Q)}(s))^T}
	\widetilde\CostN
	\nonumber\\
	&=& - (W_{Q+1}(s))^T (W_{Q+1}(s) B^{(Q)}(s) + Y) \,,
\eeqn 
and
\eqn\label{eq-derW-GD-1} 
	\partial_s W_{Q+1}(s) &=& -  \partial_{(W_{Q+1}(s))^T}
	\widetilde\CostN 
	\nonumber\\
	&=&- (W_{Q+1}(s) B^{(Q)}(s)+Y)(B^{(Q)}(s))^T \,.
\eeqn 
Next, we analyze the solutions to these gradient descent equations.

\subsection{Propagators}
We straightforwardly deduce from \eqref{eq-derB-GD-1} and \eqref{eq-derW-GD-1} that
\eqn\label{eq-ders-WB-Y-main-1} 
	\lefteqn{
	\partial_s(W_{Q+1}(s)B^{(Q)}(s)+Y)
	}
	\nonumber\\
	&&=
	- \underbrace{W_{Q+1}(s)(W_{Q+1}(s))^T }_{\geq 0}(W_{Q+1}(s) B^{(Q)}(s) + Y) 
	\nonumber\\
	&&\hspace{2cm}- (W_{Q+1}(s) B^{(Q)}(s)+Y)
	\underbrace{(B^{(Q)}(s))^T B^{(Q)}(s)}_{\geq 0} \,.
\eeqn 
We define the propagators for $s>s_0$,
\eqn\label{eq-cUB-cUW-def-1} 
	\partial_s \cU_B(s,s_0) &=& -\cU_B(s,s_0) \,  B^{(Q)}(s)(B^{(Q)}(s))^T
	\nonumber\\
	\partial_s \cU_W(s,s_0) &=& -(W_{Q+1}(s))^T W_{Q+1}(s)\, \cU_W(s,s_0) 
\eeqn 
and
\eqn\label{eq-cUB-cUW-def-1} 
	\partial_{s_0} \cU_B(s,s_0) &=& - B^{(Q)}(s_0)(B^{(Q)}(s_0))^T 
	\, \cU_B(s,s_0) \, 
	\nonumber\\
	\partial_{s_0} \cU_W(s,s_0) &=& - \cU_W(s,s_0) 
	\,(W_{Q+1}(s_0))^T W_{Q+1}(s_0) \,
\eeqn 
with $\cU_B(s_0,s_0)=\1=\cU_W(s_0,s_0)$, and find that
\eqn 
	\lefteqn{
	W_{Q+1}(s)B^{(Q)}(s)+Y
	}
	\nonumber\\
	&&\hspace{1cm}
	= \cU_W(s)\,(W_{Q+1}(0)B^{(Q)}(0)+Y)\, \cU_B(s) \,.
\eeqn  
For brevity, we write
\eqn 
	\cU_B(s):=\cU_B(s,0)
	\;\;\;{\rm and}\;\;\;
	\cU_W(s):=\cU_W(s,0) \,.
\eeqn 
Then, the following basic fact holds.

\begin{lemma}
	Assume that there exists $\lambda_0>0$ such that $B^{(Q)}(s)(B^{(Q)}(s))^T >\lambda_0$ for all $s$. Then, the operator norm bound 
	\eqn 
		\|\cU_B(s,s_0)\|_{op} \leq e^{-(s-s_0) \lambda_0 }
	\eeqn 
	holds for $s>s_0$. 
	An analogous statement is true for $\cU_W(s,s_0)$ if there exists $\lambda_0>0$ such that $(W_{Q+1}(s))^T W_{Q+1}(s)>\lambda_0$ for all $s$.
\end{lemma}

\prf
For any vector $v\in\R^Q$, we have that
\eqn
	\partial_s| \, \cU_B^T(s,s_0)v \, |^2 
	&=&-2
	\langle v, \cU_B(s,s_0)B^{(Q)}(s)(B^{(Q)}(s))^T\cU_B(s,s_0)\rangle 
	\nonumber\\
	&<&-2\lambda_0 | \, \cU_B^T(s,s_0)v \, |^2 
\eeqn 
and hence,
\eqn 
	| \, \cU_B^T(s,s_0)v \, |^2 < e^{-2(s-s_0)\lambda_0}| \, \cU_B^T(s_0,s_0)v \, |^2
	= e^{-2(s-s_0)\lambda_0} |v|^2 \,.
\eeqn 
Since this holds for arbitrary $v\in\R^Q$, and $\|\cU_B(s,s_0)\|_{op} =\|\cU_B^T(s,s_0)\|_{op} $, the claim follows.
\endprf

We note that formally, we can represent the solutions to \eqref{eq-derB-GD-1} and \eqref{eq-derW-GD-1} by use of the Duhamel (variation of constants) formula,
\eqn\label{eq-derB-GD-sol-1}
	B^{(Q)}(s) =  \cU_W(s) B^{(Q)}(0) +
	\int_0^s ds' \, \cU_W(s,s') (W_{Q+1}(s'))^T Y \,,
\eeqn
and
\eqn\label{eq-derW-GD-sol-1}
	W_{Q+1}(s) = W_{Q+1}(0) \, \cU_B(s)+
	\int_0^s ds' \, Y(B^{(Q)}(s'))^T \cU_B(s,s') \,.
\eeqn 
The combination of \eqref{eq-derB-GD-sol-1} and \eqref{eq-derW-GD-sol-1} defines a system of fixed point equations for the solution of \eqref{eq-derB-GD-1} and \eqref{eq-derW-GD-1}. However, instead of directly solving this fixed point problem, we will use the following route via a matrix-valued conservation law.

As a preparation for the subsequent discussion, we also note the following basic fact.

\begin{lemma}\label{lm-AB-lowbd-1}
	Given any $A,B\in\R^{Q\times Q}$ with  $BB^T>\lambda_0>0$, it follows that 
	\eqn 
		\|AB\|_{op} > \sqrt{\lambda_0} \|A\|_{op}
	\eeqn 
	in operator norm.
\end{lemma}

\prf
We have 
\eqn 
	\|AB\|_{op}^2 &=& \|B^TA^T\|_{op}
	\nonumber\\
	&=&\sup_{v\in\R^Q,|v|=1}\langle v, ABB^T A^T v\rangle
	\nonumber\\
	&>&\lambda_0 \sup_{v\in\R^Q,|v|=1}| A^T v|^2
	\nonumber\\
	&=&\lambda_0\|A^T\|_{op}^2 = \lambda_0 \|A\|_{op}^2 \,,
\eeqn 
as claimed.
\endprf

\subsection{Conservation laws and spectral gap}

We will derive an a priori spectral gap condition based on the existence of a conservation law along orbits of the gradient flow.
Its existence is a consequence of the fact that in order to minimize the cost, the gradient flow has to accomplish that 
\eqn 
	W_{Q+1}(s)B^{(Q)}(s)\rightarrow -Y
	\;\;\;(s\rightarrow\infty) \,.
\eeqn 
However, both $W_{Q+1}(s)$ and $B^{(Q)}(s)$ are unknowns, hence this problem is overdetermined, and \eqref{prp-conservation-1} accounts for the existence of a "gauge freedom" by which $B^{(Q)}\rightarrow A B^{(Q)}$ and $W_{Q+1}\rightarrow W_{Q+1} A^{-1}$ will yield the same result for any $A=A(s)$ with $A:\R_+\rightarrow GL(Q)$.

\begin{proposition}\label{prp-conservation-1}
	The matrix-valued integral of motion
	\eqn\label{eq-cI-conserv-def-1}
		\cI(s):=B^{(Q)}(s)(B^{(Q)}(s))^T-(W_{Q+1}(s))^T W_{Q+1}(s) 
		\;\;\;\in\R^{Q\times Q}
	\eeqn 
	is conserved along orbits of the gradient flow, that is, $\cI(s)=\cI(0)$ for all $s\in\R_+$.
\end{proposition} 

\prf
First, we note that multiplying \eqref{eq-derB-GD-1} with $(B^{(Q)}(s))^T$ from the right, and \eqref{eq-derW-GD-1} with $(W_{Q+1}(s))^T$ from the left, and subtracting, we obtain
\eqn 
	\partial_s B^{(Q)}(s)(B^{(Q)}(s))^T-(W_{Q+1}(s))^T \partial_s W_{Q+1}(s) &=& 0
	\nonumber\\
	B^{(Q)}(s)(\partial_s B^{(Q)}(s))^T-(\partial_s W_{Q+1}(s))^T  W_{Q+1}(s) &=& 0
\eeqn 
where the second line is the transpose of the first line. Adding both lines, we find
\eqn 
	\partial_s\cI(s) = 0 \,,
\eeqn 
as claimed.
\endprf 

This conservation law implies the existence of a spectral gap under the assumption of positive or negative definiteness of $\cI(0)$, uniformly in $s$. 

\begin{theorem}
	Assume that the initial data for the gradient flow allow for
	\eqn\label{eq-cI-pos-1} 
		\cI(0)= B^{(Q)}(0)(B^{(Q)}(0))^T -(W_{Q+1}(0))^T W_{Q+1}(0)
	\eeqn 
	to be either positive or negative definite, so that there exists $\lambda_0>0$ such that either
	\eqn\label{eq-cI0-lowbd-1}  
		\inf\spec(\cI(0))>\lambda_0
	\eeqn
	or
	\eqn
		\inf\spec(-\cI(0))>\lambda_0 \,.
	\eeqn 
	Then, 
	\eqn\label{eq-cost-expon-bd-1}
		\widetilde\CostN\big|_s = 
		\frac12\tr\Big((W_{Q+1}B^{(Q)}+Y)^T(W_{Q+1}B^{(Q)}+Y)\Big)\Big|_s
		\leq e^{-2s\lambda_0}
		\widetilde\CostN\big|_{s=0} \,.
	\eeqn 
	That is, the loss converges exponentially to zero as $s\rightarrow\infty$, and  
	\eqn 
		\lim_{s\rightarrow\infty}W_{Q+1}(s)B^{(Q)}(s)=-Y 
	\eeqn 
	strongly, in Hilbert-Schmidt norm.
	
	Moreover, both $\|W_{Q+1}(s)\|_{op}$ and $\|B^{(Q)}(s)\|_{op}$ are bounded, uniformly in $s$, and the limits
	\eqn 
		B^{(Q)}_\infty := \lim_{s\rightarrow\infty}B^{(Q)}(s)
		\;\;\;{\rm and} \;\;\;
		W_{Q+1,\infty} := \lim_{s\rightarrow\infty}W_{Q+1}(s)
	\eeqn 
	exist. 	
\end{theorem}

\prf 
We recall from \eqref{eq-cI-conserv-def-1} that
\eqn\label{eq-cI-conserv-def-1-1}
	\underbrace{B^{(Q)}(s)(B^{(Q)}(s))^T }_{\geq0}= \cI(0) 
	+ \underbrace{(W_{Q+1}(s))^T W_{Q+1}(s)}_{\geq 0} \,.
\eeqn 
Therefore, we find a spectral gap, uniformly in $s\in\R_+$, either given by
\eqn\label{eq-BTB-specgap-1-1}
	\inf\spec \Big((B^{(Q)}(s))^TB^{(Q)}(s)\Big)\geq \inf\spec(\cI(0)) =\lambda_0>0 
\eeqn 
if $\cI(0) >\lambda_0$ is positive definite, or
\eqn\label{eq-BTB-specgap-1-2}
	\inf\spec \Big((W_{Q+1}(s))^T W_{Q+1}(s)\Big)\geq \inf\spec(-\cI(0)) =\lambda_0>0 
\eeqn 
if $\cI(0) <-\lambda_0$ is negative definite.
Here we recalled the elementary fact that for any square matrix $A$, the spectra of $AA^T$ and $A^TA$ coincide (given an eigenvalue $\lambda$ and eigenvector $v$ with $A^TA v=\lambda v$, it follows that $A A^T(Av)=\lambda (Av)$).

Moreover, from \eqref{eq-ders-WB-Y-main-1}, we obtain
\eqn 
	\lefteqn{\partial_s \frac12\tr\Big((W_{Q+1}B^{(Q)}+Y)^T(W_{Q+1}B^{(Q)}+Y)\Big)\Big|_s
	}
	\nonumber\\
	&=& 
	- \frac12\tr\Big((W_{Q+1}B^{(Q)}+Y)^T(W_{Q+1}B^{(Q)}(s)+Y)B^{(Q)}(B^{(Q)})^T\Big)\Big|_s
	\nonumber\\
	&& 
	- \frac12\tr\Big(B^{(Q)}(B^{(Q)})^T(W_{Q+1}B^{(Q)}+Y)^T(W_{Q+1}B^{(Q)}(s)+Y)\Big)\Big|_s
	\nonumber\\
	&&
	\hspace{1cm}
	- \tr\Big((W_{Q+1}B^{(Q)}+Y)^T W_{Q+1}W_{Q+1}^T  (W_{Q+1}B^{(Q)}+Y)\Big)\Big|_s
	\nonumber\\
	&=& 
	- \tr\Big((W_{Q+1}B^{(Q)}(s)+Y)B^{(Q)}(B^{(Q)})^T(W_{Q+1}B^{(Q)}+Y)^T\Big)\Big|_s
	\nonumber\\
	&&\hspace{1cm}
	- \tr\Big((W_{Q+1}B^{(Q)}+Y)^T W_{Q+1} W_{Q+1}^T (W_{Q+1}B^{(Q)}+Y)\Big)\Big|_s
	\nonumber\\
	&\leq&-\big(\underbrace{
	\inf\spec (B^{(Q)}(B^{(Q)})^T)+\inf\spec(W_{Q+1} W_{Q+1}^T)
	}_{\geq\lambda_0 \;{\rm uniformly \; in \;}s}\big)\Big|_s 
	\nonumber\\
	&&\hspace{2cm}
	\tr\Big((W_{Q+1}B^{(Q)}+Y)^T(W_{Q+1}B^{(Q)}+Y)\Big)\Big|_s
	\nonumber\\
	&\leq&-\lambda_0 \tr\Big((W_{Q+1}B^{(Q)}+Y)^T(W_{Q+1}B^{(Q)}+Y)\Big)\Big|_s
\eeqn 
where we used cyclicity of the trace, especially with $\tr(A^TA)=\tr(AA^T)$, and the spectral gap condition \eqref{eq-BTB-specgap-1-1}, respectively \eqref{eq-BTB-specgap-1-2}. 
Integrating with respect to $s$, we arrive at \eqref{eq-cost-expon-bd-1}.

To prove that $\| (W_{Q+1}(s) \|_{op}$ and $\|B^{(Q)}(s)\|_{op}$ are uniformly bounded in $s$, we first assume the case \eqref{eq-BTB-specgap-1-1}. Then,
\eqn 
	\lefteqn
	{
	\lambda_0 \| (W_{Q+1}(s)+Y(B^{(Q)}(s))^{-1})    \|_{op}^2
	}
	\nonumber\\
	&\leq&\| (W_{Q+1}(s)+Y(B^{(Q)}(s))^{-1})B^{(Q)}(s)   \|_{op}^2
	\nonumber\\
	&=&\| (W_{Q+1}(s)B^{(Q)}(s)+Y )   \|_{op}^2
	\nonumber\\
	&\leq& \tr\Big(\Big|W_{Q+1}(s)B^{(Q)}(s)+Y \Big|^2\Big)
	\nonumber\\
	&<&c_{\widetilde\CostN} e^{-2s\lambda_0} \,.
\eeqn 
where we used Lemma \ref{lm-AB-lowbd-1} in the first step, and \eqref{eq-cost-expon-bd-1} in the last step. The constant $c_{\widetilde\CostN} $ is proportional to the standard loss at $s=0$.
This implies that 
\eqn 
	\| W_{Q+1}(s) \|_{op}
	&\leq& \|Y(B^{(Q)}(s))^{-1})    \|_{op} + \frac{c_{\widetilde\CostN} }{\sqrt{\lambda_0}} e^{-2s\lambda_0}
	\nonumber\\
	&\leq&\frac{1}{\sqrt{\lambda_0}}\Big(\|Y\|_{op}+ c_{\widetilde\CostN}  e^{-2s\lambda_0} \Big)
\eeqn 
where we used \eqref{eq-BTB-specgap-1-1} in the last step. The right hand side is bounded, uniformly in $s$, therefore there exists a constant $c_W$ such that
\eqn 
	\| W_{Q+1}(s) \|_{op} < c_W \,.
\eeqn 
But then, the conservation law \eqref{eq-cI-conserv-def-1-1} implies that
\eqn 
	\|B^{(Q)}(s)\|_{op}^2< \|\cI(0)\|_{op} + c_W^2 =: c_B^2
\eeqn 
is uniformly bounded in $s$.

Moreover, we obtain from \eqref{eq-derW-GD-1} that
\eqn\label{eq-derW-GD-1-1} 
	\|\partial_s W_{Q+1}(s)\|_{op} &\leq&
	\|(W_{Q+1}(s) B^{(Q)}(s)+Y)\|_{op}\|B^{(Q)}(s)\|_{op}
	\nonumber\\
	&\leq&c_{\widetilde\CostN}  c_B e^{-s\lambda_0} 
\eeqn 
converges to zero as $s\rightarrow\infty$. Therefore, $\lim_{s\rightarrow\infty}W_{Q+1}(s)$ exists, and the convergence is exponential. 
On the other hand, \eqref{eq-derB-GD-1} implies that
\eqn\label{eq-derB-GD-1-1}
	\|\partial_s B^{(Q)}(s)\|_{op} &\leq&  
	\|W_{Q+1}(s) B^{(Q)}(s) + Y\|_{op} \|W_{Q+1}(s)\|_{op}
	\nonumber\\
	&\leq&c_{\widetilde\CostN}  c_W e^{-s\lambda_0} \,.
\eeqn 
Therefore, $\lim_{s\rightarrow\infty}B^{(Q)}(s)$ also exists, with exponential convergence rate. 

The proof for the case \eqref{eq-BTB-specgap-1-2} is similar.
\endprf


\section{Standard loss and clustered initial data}
\label{sec-std-clustered-1-0-0}

Next, we consider another situation in which the initial data for the cumulative weights and biases yield a particularly simple structure, which allows for the explicit solution of the gradient flow equations in the output layer. Namely, we choose initial data $(W^{(\ell)}(s=0),\beta^{(\ell)}(s=0))_\ell$ in a manner that
\eqn 	
	\tau^{(Q)}\big|_{s=0}(x^{(0)}_{\ell,i}) = x^{(0)}_{\ell,i}
	\;\;\;,\;
	i_\ell=1,\dots,N_\ell 
	\;\;\;,\;
	\ell=1,\dots,Q \,.
\eeqn 	
That is, the $\ell$-th cluster is mapped to itself, for every $\ell=1,\dots,Q$.
The explicit construction of initial data satisfying these properties is presented in \cite{cheewa-2,cheewa-4}; it corresponds to the case in which every cluster is located in the positive sector of every truncation map, that is,
\eqn
	x_{\ell,i}^{(0)} \in \cS_{\ell'}^+
	\;\;\;{\rm for\;all}\;
	\ell,\ell'=1,\dots,Q
	\;\;,\;
	i=1,\dots,N_\ell\,.
\eeqn 
In this situation, 
\eqn 
	\widetilde\CostN\big|_{s=0} 
	&=&\frac1{2N} \sum_{\ell=1}^Q \sum_{i=1}^{N_\ell }
	|W_{Q+1}x^{(0)}_{\ell,i}-y_\ell|^2
	\nonumber\\
	&=&\frac1{2N}\tr\Big(|W_{Q+1}X^{(0)}-Y_{ext}|^2\Big)
\eeqn 
where
\eqn 
	X^{(0)}:=[x^{(0)}_{1,1}\cdots x^{(0)}_{\ell,i_\ell}\cdots x^{(0)}_{Q,N_Q}]
	\in\R^{Q\times N}
\eeqn 
and
\eqn 
	Y_{ext}:=[y_1\cdots  
	\underbrace{y_\ell\cdots y_\ell }_{N_\ell\;{\rm copies}}\cdots   y_Q]
	\in\R^{Q\times N} \,.
\eeqn 
In particular, the loss does not depend on the cumulative parameters $(W^{(\ell)}(s=0),\beta^{(\ell)}(s=0))_\ell$, hence,
\eqn
	\partial_s W^{(\ell)}(s) = 0
	\;\;\;{\rm and}\;\;\;
	\partial_s \beta^{(\ell)}(s) = 0
\eeqn 
for all $\ell=1,\dots,Q$.
Then, the gradient flow for $W_{Q+1}$ is given by
\eqn\label{eq-derW-triv-GF-1}
	\partial_s W_{Q+1}(s) = - \frac1N(W_{Q+1}X^{(0)}-Y_{ext})(X^{(0)})^T \,.
\eeqn 
Similarly as in \eqref{eq-derW-GD-sol-1}, the solution can be represented via the Duhamel formula as
\eqn\label{eq-derW-triv-sol-1} 
	W_{Q+1}(s) &=& W_{Q+1}(0) \, e^{-\frac sN X^{(0)}(X^{(0)})^T}
	\nonumber\\
	&&\hspace{1cm}+
	\frac1N \int_0^s ds'\, Y(X^{(0)})^T e^{-\frac{(s-s')}{N} X^{(0)}(X^{(0)})^T} \,.
\eeqn 
We thus obtain the following result.

\begin{theorem}
Assume that
\eqn 
	X^{(0)}(X^{(0)})^T \in GL(Q)
\eeqn 
is invertible. 
	Then, the gradient flow \eqref{eq-derW-triv-GF-1} has the explicit solution
\eqn 
	W_{Q+1}(s) = W_{Q+1}(0) \, e^{-\frac{s}{N} X^{(0)}(X^{(0)})^T}+
	Y(X^{(0)})^+(1-e^{-\frac{s}{N} X^{(0)}(X^{(0)})^T})
\eeqn 	
where 
\eqn 
	(X^{(0)})^+:= (X^{(0)})^T(X^{(0)}(X^{(0)})^T)^{-1}
\eeqn 
is the generalized inverse of $X^{(0)}$.
Therefore, 
\eqn\label{eq-Wlim-triv-1} 
	\lim_{s\rightarrow\infty}W_{Q+1}(s) =
	Y(X^{(0)})^+ \,.
\eeqn 
\end{theorem}

Notably, \eqref{eq-Wlim-triv-1} corresponds to the loss minimizer obtained in \cite{cheewa-1} for the scenario at hand.

\appendix 

\section{Proof of Theorem \ref{thm-gradflow-input-1-0}}
\label{sec-prf-thm-gradflow-input-1-0}

Here we prove Theorem \ref{thm-gradflow-input-1-0}.
By Definition \ref{def-Wsig-aligned-1-0}, the assumption that $W^{(\ell)}$
are aligned with $\sigma$, means that in the polar decomposition
\eqn 
	W^{(\ell)} = W^{(\ell)}_*  R_\ell
\eeqn 
with $W^{(\ell)}_* = |W^{(\ell)}|$ and $R_\ell\in O(Q)$, the matrix $W^{(\ell)}_*$ is diagonal, for $\ell=1,\dots,Q$. 

It then follows that
\eqn 
	\tau^{(\ell)}(x)&=& \tau_{W^{(\ell)},\beta^{(\ell)}}(x)
	\nonumber\\
	&=& 
	R_\ell^T (W^{(\ell)}_*)^{-1}\sigma(W^{(\ell)}_* R(x+\beta^{(\ell)})) - \beta^{(\ell)}
	\nonumber\\
	&=& R_\ell^T  \sigma( R_\ell(x+\beta^{(\ell)})) - \beta^{(\ell)}
	\nonumber\\
	&=&
	\tau_{R_\ell,\beta^{(\ell)}}(x)
\eeqn 
because $\sigma(Dx)=D\sigma(x)$ for any positive semidefinite diagonal matrix $D\in\R^{Q\times Q}$.
Therefore, $\tau^{(\ell)}$ is independent of $W^{(\ell)}_*$ when it is diagonal. Thus, if $W^{(\ell)}$ and $\sigma$ are aligned, we may assume 
\eqn\label{eq-Well-Rell-1-0}
	W^{(\ell)} = R_\ell \in O(Q)\,.
\eeqn 
without any loss of generality.

For completeness, we determine the expression for the gradient with respect to $\beta^{\ell}$ for general $W^{(\ell)}$ at first,
\eqn 
	\partial_{\beta^{(\ell)}} \CostN 
	&=& \partial_{\beta^{(\ell)}} \sum_{\ell=1}^Q
	\frac12\int dx\, \mu_\ell(x)
	\Big|\tau^{(Q)}(x)-\widetilde y_\ell \Big|^2
	\nonumber\\
	&=& 
	\int dx \, \mu_\ell(x)
	(\partial_{\beta^{(\ell)}} \tau^{(\ell)}(x))^T
	(\tau^{(\ell)}(x)-\widetilde y_\ell)
	\nonumber\\
	&=& - 
	\int dx \, \mu_\ell(x)
	((W^{(\ell)})^{-1} H^{(\ell) \perp}(x)W^{(\ell)})^T
	(\tau^{(\ell)}(x)-\widetilde y_\ell)
	\nonumber\\
	&=& - 
	\int dx \, \mu_\ell(x)
	(W^{(\ell)})^{T} H^{(\ell) \perp}(x)
	(W^{(\ell)})^{-T}
	(\tau^{(\ell)}(x)-\widetilde y_\ell) \,.
\eeqn 
Hence, we may focus on the single entry $\tau^{(\ell)}(x)$, for each given $\ell\in\{1,\dots,Q\}$.

Given \eqref{eq-Well-Rell-1-0}, the above then reduces to
\eqn 
	\partial_{\beta^{(\ell)}} \CostN 
	&=& - 
	\int dx \, \mu_\ell(x)
	R_\ell^T   H^{(\ell) \perp}(x)
	 R_\ell 
	\nonumber\\
	&&\hspace{2cm}
	(R_\ell^T (\sigma(R_\ell(x+\beta^{(\ell)}))
	-\beta^{(\ell)}-\widetilde y_\ell)
\eeqn 
because 
\eqn
	W_*^{(\ell)} H^{(\ell) \perp}(x)
	(W_*^{(\ell)})^{-1} = H^{(\ell) \perp }(x)
\eeqn 
since both $W_*^{(\ell)}$ and $H^{(\ell) \perp}(x_{\ell,i}^{(0)})$ are diagonal matrices.

This further reduces to 
\eqn\label{eq-partbeta-prf-1-1-0}
	\partial_{\beta^{(\ell)}} \CostN 
	&=&  - 
	\int dx \, \mu_\ell(x)
	R_\ell^T   H^{(\ell) \perp}(x) R_\ell 
	\nonumber\\
	&&\hspace{2cm}
	(R_\ell^T  \sigma( R_\ell(x+\beta^{(\ell)}))
	-\beta^{(\ell)}-\widetilde y_\ell)
	\nonumber\\
	&=& 
	\int dx\,\mu_\ell(x)
	R_\ell^T   H^{(\ell) \perp}(x)
	 R_\ell  
	( \beta^{(\ell)}+\widetilde y_\ell)
\eeqn 
because of the key cancelation property
\eqn
	H^{(\ell) \perp}(x) \sigma( R_\ell(x+\beta^{(\ell)})) = 0 \,,
\eeqn 
which follows from orthogonality due to the use of the Euclidean metric in the definition of the loss \eqref{eq-CostNEucl-def-1-0} in input space. It does not in general hold for the standard loss \eqref{eq-CostNpullb-def-1-0} formulated with the pullback metric.

Only the term $H^{(\ell) \perp}(x_{\ell,i}^{(0)})$ depends on the training data, and we obtain
\eqn 
	\partial_{\beta^{(\ell)}} \CostN 
	&=&  
	R_\ell^T \Big(
	\int dx \, \mu_\ell(x)
	H^{(\ell) \perp}(x)\Big)
	R_\ell  
	( \beta^{(\ell)}+\widetilde y_\ell)
	\nonumber\\
	&=&  
	R_\ell^T \Big(
	\int dx \, \mu_\ell(R_\ell^T x-\beta^{(\ell)})
	H^{\perp}(x)\Big)
	R_\ell  
	( \beta^{(\ell)}+\widetilde y_\ell)
\eeqn 
where we used the coordinate transformation $x\rightarrow R_\ell x -\beta^{(\ell)}=a_{R_\ell,\beta^{(\ell)}}^{-1}(x)$ to pass to the second line. This is the asserted result.

Next, we focus on the gradient flow equations for $R_\ell\in O(Q)$. To begin with, we note that  
\eqn 
	(\partial_s R_\ell(s)) R_\ell^T(s) \in o(Q) 
\eeqn
is given by the restriction of $-\partial_{R_\ell}\CostN$ to $o(Q)$.
To find the latter, we use the following lemma.

\begin{lemma}
\label{lm-antisym-1-0}
	Let $R(s)=\exp(s\omega)\in O(Q)$ be an orbit parametrized by $s\in\R$, and with generator $\omega=-\omega^T\in o(Q)$. Then, for any smooth $f: O(Q)\rightarrow\R$,
	\eqn 
		\partial_s f(R(s)) = 
		-\tr\Big(\omega \;\pi_-\big(\partial_R f(R(s))R^T(s)\big)\Big) \,.
	\eeqn 
	Therefore, the gradient of $f(R)$, restricted to $o(Q)$, is given by 
	$\pi_-\big(\partial_R f(R)R^T\big)$.
\end{lemma}

\prf
We have 
\eqn 
	\partial_s f(R(s)) &=& \sum_{ij}\partial_{R_{ij}}f(R(s))\partial_s R_ij(s)
	\nonumber\\
	&=&\tr\Big((\partial_R f(R(s))^T \partial_s R(s) \Big)
	\nonumber\\
	&=&\tr\Big((\partial_R f(R(s))^T \omega R(s) \Big)
	\nonumber\\
	&=&-\tr\Big(\omega(\partial_R f(R(s)) R^T(s) \Big)
\eeqn 
using that $\tr(AB)=\tr(B^TA^T)$, cyclicity of the trace, and antisymmetry of the generator, $\omega^T=-\omega$, to pass to the last line.
Since $\tr(\omega A)=0$ for all symmetric $A=A^T\in\R^{Q\times Q}$, it follows that $\tr(\omega A)=\tr(\omega \pi_-(A))$ for all $A\in\R^{Q\times Q}$, and we arrive at the claim.
\endprf

To obtain the gradient for the Euclidean loss in input space, we first determine
\eqn 
	\lefteqn{
	\partial_{R_{jk}}(\tau_{R,\beta}(x)-\widetilde y)_i
	}
	\nonumber\\
	&=&
	\partial_{R_{jk}}\Big(\sum_r 
	R_{ri}\sigma\big(\sum_{s} R_{rs}(x_s+\beta_s)\big)
	-(\beta_i+\widetilde y_i)\Big) 
	\nonumber\\
	&=& 
	\sum_r \delta_{rj} \delta_{ki} \sigma\big(\sum_{s} R_{rs}(x_s+\beta_s)\big)
	+
	\sum_r R_{ri}\delta_{jr}h\Big(\sum_{s}R_{is}(x_s+\beta_s)\Big)(x_k+\beta_k) 	
	\nonumber\\
	&=&
	\delta_{ki}\sigma\big(\sum_{s} R_{js}(x_s+\beta_s)
	+
	R_{ji} (h(R(x+\beta)))_j (x_k+\beta_k) \,.
\eeqn 
Therefore,
\eqn 
	\lefteqn{
	\partial_{R_{jk}}\frac12|\tau_{R,\beta}(x)-\widetilde y|^2
	}
	\nonumber\\
	&=&
	\partial_{R_{jk}}\frac12\sum_i(\tau_{R,\beta}(x)-\widetilde y)_i^2
	\nonumber\\
	&=&
	\sum_i
	\Big(\delta_{ki}\sigma\big(\sum_{s} R_{js}(x_s+\beta_s)
	+
	R_{ji} (h(R(x+\beta)))_j (x_k+\beta_k)
	\Big)
	(\tau_{R,\beta}(x)-\widetilde y)_i
	\nonumber\\
	&=&
	\sigma(R(x+\beta))_j
	(\tau_{R,\beta}(x)-\widetilde y)_k
	\nonumber\\
	&&\hspace{2cm}+
	(h(R(x+\beta)))_j( R 
	(\tau_{R,\beta}(x)-\widetilde y))_j (x_k+\beta_k)
	\Big)
	\nonumber\\
	&=&
	(h(R(x+\beta)))_j(R(x+\beta))_j
	(\tau_{R,\beta}(x)-\widetilde y)_k
	\nonumber\\
	&&\hspace{2cm}+
	(h(R(x+\beta)))_j( R 
	(\tau_{R,\beta}(x)-\widetilde y))_j (x_k+\beta_k)
\eeqn 
using $\sigma(x)=H(x)x$ for $x\in\R^Q$, and in matrix notation,
\eqn\label{eq-derR-tausq-1-0}
	\lefteqn{
	[\partial_{R_{jk}}\frac12|\tau_{R,\beta}(x)-\widetilde y|^2]_{jk}
	}
	\nonumber\\
	&=&
	H_{R,\beta}(x)R(x+\beta)(\tau_{R,\beta}(x)-\widetilde y)^T
	\nonumber\\
	&&
	\hspace{2cm}
	+H_{R,\beta}(x)R(\tau_{R,\beta}(x)-\widetilde y) 
	\; (x+\beta)^T \,.
\eeqn
Thus, we obtain
\eqn  
	\lefteqn{
	\Big(\partial_{R}\frac12|\tau_{R,\beta}(x)-\widetilde y|^2\Big) R^T
	}
	\nonumber\\
	&=&
	H_{R,\beta}(x)R(x+\beta)(\tau_{R,\beta}(x)-\widetilde y)^T R^T
	\nonumber\\
	&&
	\hspace{2cm}
	+H_{R,\beta}(x)R(\tau_{R,\beta}(x)-\widetilde y) 
	\; (x+\beta)^T R^T
	\nonumber\\
	&=&
	H_{R,\beta}(x)R(x+\beta)(\sigma(R(x+\beta))-R(\beta+\widetilde y))^T 
	\nonumber\\
	&&
	\hspace{2cm}
	+H_{R,\beta}(x)(\sigma(R(x+\beta))-R(\beta+\widetilde y))  (x+\beta)^T R^T
	\nonumber\\
	&=&
	2 H_{R,\beta}(x)R(x+\beta)(x+\beta)^TR^TH_{R,\beta}(x) 
	\\
	&&
	\hspace{1.5cm}
	-H_{R,\beta}(x)R(x+\beta)(\beta+\widetilde y)^T R^T 
	-H_{R,\beta}(x)R(\beta+\widetilde y) (x+\beta)^T R^T
	\nonumber\,.
\eeqn
The gradient of the loss with respect to $R_\ell$ therefore yields
\eqn\label{eq-gradCN-b-1-0-0}
	\lefteqn{
	(\partial_{R_\ell}\CostN)R_\ell^T
	}
	\nonumber\\
	&=&
	2 \int dx\, \mu_\ell(x)
	H_{R_\ell,\beta}(x)R_\ell  
	(x+\beta^{(\ell)})
	\; (x+\beta^{(\ell)})^T R_\ell^T H_{R_\ell,\beta}(x)
	\\
	&& 
	- \int dx\, \mu_\ell(x) \, H_{R_\ell,\beta}(x)R_\ell  
	\Big( (\beta^{(\ell)}+\widetilde y) (x+\beta^{(\ell)})^T + 
	(x+\beta^{(\ell)})(\beta^{(\ell)}+\widetilde y)^T \Big) 
	R_\ell^T
	\,.
	\nonumber
\eeqn 
Applying the antisymmetrization operator $\pi_-$, the first term on the r.h.s. is eliminated, and we obtain
\eqn
	\lefteqn{
	\pi_-\Big((\partial_{R_\ell}\CostN)R_\ell^T\Big)
	}
	\nonumber\\
	&=&
	-\frac12\int dx\, \mu_\ell(x)
	[H_{R_\ell,\beta}(x) \;,\; 
	\nonumber\\
	&&\hspace{1cm}
	R_\ell \Big( (\beta^{(\ell)}+\widetilde y) (x+\beta^{(\ell)})^T + 
	(x+\beta^{(\ell)})(\beta^{(\ell)}+\widetilde y)^T \Big) 
	R_\ell^T ]
	\\ 
	&=&- \int dx\, \mu_\ell(R_\ell^T x-\beta^{(\ell)})
	[ H(x) \; , \;  
	M^{(\ell)}(x) \; ]
	\nonumber\\
	&=& 
	-\int_{\R^Q\setminus(\R_+^Q\cup \R_-^Q)} dx \, \mu_\ell(a_{R_\ell,\beta^{(\ell)}}^{-1}(x))
	\big[H(x)\;,\;
	M^{(\ell)}(x) \big]
	\,,
\eeqn 
where 
\eqn 
	M^{(\ell)}(x) = \frac12\Big(R_\ell (\beta^{(\ell)}+\widetilde y)x^T + 
	x (\beta^{(\ell)}+\widetilde y)^T R_\ell^T \Big)\,.
\eeqn 
Here we applied the coordinate transformation 
\eqn
	x\rightarrow R_\ell x-\beta^{(\ell)}=a_{R_\ell,\beta^{(\ell)}}^{-1}(x)\,,
\eeqn
and we observed that the commutator is trivially zero on $\R_+^Q\cup \R_-^Q$, because $H(x)=\1$ for all $x\in\R_+^Q$ and $H(x)=0$ for all $x\in\R_-^Q$.
Thus, we arrive at \eqref{eq-Om-commut-def-1-0}. 

Finally, we note that
\eqn
	(\partial_{\beta^{(\ell)}}\CostN) \cdot \partial_s\beta^{(\ell)}
	= - \Big|R_\ell^T \Big(
	\int dx\,\mu_\ell(x)
	  H^{(\ell) \perp}(x)\Big)
	 R_\ell  
	( \beta^{(\ell)}+\widetilde y_\ell)\Big|^2
	\;\leq \;0
\eeqn 
follows immediately from \eqref{eq-partbeta-prf-1-1-0}. Furthermore, we have
\eqn 
	\tr\Big((\partial_{R_\ell}\CostN)^T\partial_s R_\ell\Big)
	&=&\tr\Big(((\partial_{R_\ell}\CostN)R_\ell^T)^T
	((\partial_s R_\ell)R_\ell^T)\Big)
	\nonumber\\
	&=&- \tr\Big(((\partial_{R_\ell}\CostN)R_\ell^T)^T 
	\pi_- ((\partial_{R_\ell}\CostN)R_\ell^T)\Big)
	\nonumber\\
	&=&- \tr\Big((\pi_-((\partial_{R_\ell}\CostN)R_\ell^T))^T 
	\pi_- ((\partial_{R_\ell}\CostN)R_\ell^T)\Big)
	\nonumber\\
	&=& 
	- \tr\big(\;|\Omega_\ell|^2\;\big)
	\;\leq\;0
\eeqn 
using cyclicity of the trace, and the same arguments as in the proof of Lemma \ref{lm-antisym-1-0}.

This completes the proof of Theorem \ref{thm-gradflow-input-1-0}.
\qed

\section{Proof of Theorem \ref{thm-gradflow-general-1-0}}

Here we prove Theorem \ref{thm-gradflow-general-1-0}.

\begin{lemma}
\label{lm-tau-P-1-0-0}
Let for $\ell_2\geq \ell_1$, 
\eqn 
	\tau^{(\ell_1,\ell_2)}(x) := 
	\tau^{(\ell_2)}\circ\tau^{(\ell_2-1)}
	\circ\cdots\circ\tau^{(\ell_1)}(x) \,.
\eeqn 
Then, 
\eqn
	\tau^{(\ell_1,\ell_2)}(x)
	&=&P^+_{\ell_1,\ell_2}(x)x 
	-
	\sum_{\ell=\ell_1}^{\ell_2}
	P^-_{\ell,\ell_2}(x) \beta^{(\ell)}
\eeqn
where
\eqn 
	P^+_{\ell_1,\ell_2}(x)&:=& R_{\ell_2}^T H^{(\ell_2)} R_{\ell_2}
	\cdots
	R_{\ell_1}^T H^{(\ell_1)} R_{\ell_1}\,,
\eeqn  
and
\eqn 
	P^-_{\ell,\ell_2}(x) &:=& 
	R_{\ell_2}^T H^{(\ell_2)} R_{\ell_2} \cdots
	R_{\ell+1}^T H^{(\ell+1)} R_{\ell+1}
	R_{\ell}^T H^{(\ell) \perp } R_{\ell}
	\nonumber\\
	P^-_{\ell,\ell}(x) &:=& 
	R_{\ell}^T H^{(\ell) \perp} R_{\ell} \,,
\eeqn 
using  
\eqn 
	H^{(\ell)} \equiv H^{(\ell)}(\tau^{(\ell-1,1)}(x))
	=H(R_\ell(\tau^{(\ell-1,1)}(x)+\beta^{(\ell)} ) )
\eeqn
for notational brevity.
\end{lemma}

\prf
Recalling \eqref{eq-tau-H-id-1-0}, we apply 
\eqn
	\tau^{(\ell)}(x) = R_\ell^T H^{(\ell)}(x) R_\ell x 
	- R_\ell^T H^{(\ell) \perp}(x) R_\ell \beta^{(\ell)}
\eeqn
to obtain
\eqn
	\lefteqn{
	\tau^{(\ell_2,\ell_1)}(x) 
	}
	\nonumber\\
	&=& R_{\ell_2}^T H^{(\ell_2)} R_{\ell_2}  
	\tau^{(\ell_2-1,\ell_1)}(x) 
	- R_{\ell_2}^T H^{(\ell_2) \perp} R_{\ell_2} \beta^{(\ell_2)}
	\nonumber\\
	&=& 
	R_{\ell_2}^T H^{(\ell_2)} R_{\ell_2}  
	R_{\ell_2-1}^T H^{(\ell_2-1)} R_{\ell_2-1}  
	\tau^{(\ell_2-2,\ell_1)}(x) 
	\nonumber\\
	&&\hspace{1cm}
	- R_{\ell_2}^T H^{(\ell_2)} R_{\ell_2}  
	R_{\ell_2-1}^T H^{(\ell_2-1) \perp } R_{\ell_2-1}
	 \beta^{(\ell_2-1)}
	\nonumber\\
	&&\hspace{2cm}
	- R_{\ell_2}^T H^{(\ell_2) \perp} R_{\ell_2} \beta^{(\ell_2)}
\eeqn
and recursively,
\eqn
	&=&
	R_{\ell_2}^T H^{(\ell_2)} R_{\ell_2} \cdots
	R_{\ell_1}^T H^{(\ell_1)} R_{\ell_1} 
	x
	\nonumber\\
	&&\hspace{1cm}
	- R_{\ell_2}^T H^{(\ell_2)} R_{\ell_2} \cdots
	R_{\ell_1+1}^T H^{(\ell_1+1)} R_{\ell_1+1}
	R_{\ell_1}^T H^{(\ell_1) \perp } R_{\ell_1}
	 \beta^{(\ell_1)}
	\nonumber\\
	&&\hspace{2cm}
	- \cdots -
	R_{\ell_2}^T H^{(\ell_2) \perp} R_{\ell_2} \beta^{(\ell_2)}
	\nonumber\\
	&=&
	P^+_{\ell_1,\ell_2}(x)x 
	-
	\sum_{\ell=\ell_1}^{\ell_2}
	P^-_{\ell,\ell_2}(x) \beta^{(\ell)}
\eeqn 
with
\eqn 
	P^-_{\ell,\ell_2}(x) = 
	R_{\ell_2}^T H^{(\ell_2)} R_{\ell_2} \cdots
	R_{\ell+1}^T H^{(\ell+1)} R_{\ell+1}
	R_{\ell}^T H^{(\ell) \perp } R_{\ell}
\eeqn 
and
\eqn 
	P^-_{\ell_2,\ell_2}(x) = 
	R_{\ell_2}^T H^{(\ell_2) \perp} R_{\ell_2}
\eeqn 
as claimed.
\endprf

\begin{lemma}
For any $x\in\R^Q$, the following holds,
\eqn 
	\lefteqn{
	\partial_{\beta^{(\ell)}}\frac12
	\Big| \, \tau^{(Q,1)}(x)-\widetilde y \, \Big|^2
	}
	\nonumber\\
	&=&
	(P^-_{\ell,Q}(x))^T
	\Big(P^+_{\ell,Q}(x)\tau^{(\ell-1,1)}(x)
	- \sum_{\ell'=\ell}^{Q}
	P^-_{\ell',Q}(x) \beta^{(\ell')}
	-\widetilde y\Big) \,,
\eeqn
using the same notations as in Lemma \ref{lm-tau-P-1-0-0}.
\end{lemma}

\prf
Recalling the abbreviated notation
\eqn 
	H^{(\ell')}\equiv H^{(\ell')}(\tau^{(\ell',1)}(x)) \,,
\eeqn 
we determine, for $1\leq \ell \leq Q$,
\eqn 
	\lefteqn{
	\partial_{\beta^{(\ell)}}\tau^{(Q,1)}(x)
	}
	\nonumber\\
	&=&(\partial_{x'}\tau^{(Q)}(x'))\big|_{x'=\tau^{(Q-1,1)}(x)}
	\circ
	\cdots\circ 
	(\partial_{x'}\tau^{(\ell+1)}(x'))\big|_{x'=\tau^{(\ell,1)}(x)}
	\nonumber\\
	&&	
	\hspace{2cm}
	\circ \; \partial_{\beta^{(\ell)}}\tau^{(\ell)}
	\Big(\tau^{(\ell-1,1)}(x)\Big)
	\nonumber\\
	&=&R_Q^T H^{(Q)} R_Q
	\cdots 
	\cdots R_{\ell+1}^T H^{(\ell+1)} R_{\ell+1}
	\big(-R_\ell^T H^{(\ell) \perp} R_{\ell}\big)
	\nonumber\\
	&=&-P^-_{\ell,Q}(x) \,.
\eeqn 
We find, for $\widetilde y\in \R^Q$,
\eqn 
	\lefteqn{
	\partial_{\beta^{(\ell)}}\frac12
	\Big| \, \tau^{(Q,1)}(x)-\widetilde y \, \Big|^2
	}
	\nonumber\\
	&=&
	(\partial_{\beta^{(\ell)}}\tau^{(Q,1)}(x))^T 
	(\tau^{(Q,1)}(x)-\widetilde y)
	\nonumber\\
	&=&
	-(P^-_{\ell,Q}(x))^T
	\Big(P^+_{\ell,Q}(x)\tau^{(\ell-1,1)}(x)
	- \sum_{\ell'=\ell}^{Q}
	P^-_{\ell',Q}(x) \beta^{(\ell')}
	-\widetilde y\Big) \,,
\eeqn 
as claimed.
\endprf

As a consequence, we obtain
\eqn  
	\partial_{\beta^{(\ell)}}\CostN 
	&=&\sum_{\ell'=1}^Q
	\int dx \, \mu_{\ell'}(x)
	\frac12
	\partial_{\beta^{(\ell)}}
	\Big| \, \tau^{(Q,1)}(x)-\widetilde y \, \Big|^2
	\nonumber\\
	&=&
	-\sum_{\ell'=\ell}^Q
	\int dx \, \mu_{\ell'}(x)
	(P^-_{\ell',Q}(x))^T
	\Big(P^+_{\ell',Q}(x)\tau^{(\ell'-1,1)}(x)
	\nonumber\\
	&&\hspace{1cm}
	- \sum_{\ell''=\ell'}^{Q}
	P^-_{\ell'',Q}(x) \beta^{(\ell'')}
	-\widetilde y\Big) \,,
\eeqn
which yields the asserted expression for $\partial_s\beta^{(\ell)}$.

\begin{lemma}
For any $x\in\R^Q$, 
\eqn 
	\lefteqn{
	\partial_{R_\ell}\frac12
	\Big| \, \tau^{(Q,1)}(x)-\widetilde y_{\ell'} \, \Big|^2
	}
	\nonumber\\ 
	&=&
	H^{(\ell)} R_{\ell}(\tau^{(\ell-1,1)}(x)+\beta^{(\ell)})
	((P^+_{\ell+1,Q})^T(\tau^{(Q,1)}(x)-\widetilde y_{\ell'}))^T 
	\nonumber\\
	&& 
	+H^{(\ell)} 
	R_{\ell}
	(P^+_{\ell+1,Q})^T(\tau^{(Q,1)}(x)-\widetilde y_{\ell'})
	(\tau^{(\ell-1,1)}(x)+\beta^{(\ell)})^T \,.
\eeqn
using the same notations as in Lemma \ref{lm-tau-P-1-0-0}.
\end{lemma}

\prf
To begin with, we have
\eqn 
	\lefteqn{
	\partial_{R_{jk}}(\tau_{R,\beta}(x)-\widetilde y)_i
	}
	\nonumber\\
	&=&
	\partial_{R_{jk}} \sum_{r} \Big(R_{ri}
	\sigma(\sum_{s}  R_{rs}(x_s+\beta_s)) -\beta_i\Big)
	\nonumber\\
	&=&
	\sum_{r}\Big( \delta_{jr}\delta_{ki}
	\sigma(\sum_{s} R_{rs}(x_s+\beta_s))
	+\delta_{jr} R_{ri}
	h(\sum_{s}  R_{rs}(x_s+\beta_s))(x_k+\beta_k)\Big)
	\nonumber\\
	&=&
	\delta_{ki}
	\sigma(\sum_s R_{js}(x_s+\beta_s))
	+R_{ji}h(\sum_s R_{js}(x_s+\beta_s))(x_k+\beta_k) \,.
\eeqn 
Next, for $1\leq \ell \leq Q$,
\eqn 
	\lefteqn{
	\partial_{(R_\ell)_{jk}}\tau^{(Q,1)}(x)
	}
	\nonumber\\
	&=&(\partial_{x'}\tau^{(Q)}(x'))\big|_{x'=\tau^{(Q-1,1)}(x)}
	\circ
	\cdots\circ 
	(\partial_{x'}\tau^{(\ell+1)}(x'))\big|_{x'=\tau^{(\ell,1)}(x)}
	\nonumber\\
	&&	
	\hspace{2cm}
	\circ \; \partial_{(R_\ell)_{jk}}\tau^{(\ell)}
	\Big(\tau^{(\ell-1,1)}(x)\Big)
	\nonumber\\
	&=&R_Q^T H^{(Q)}(\tau^{(Q-1,1)}(x)) R_Q
	\cdots
	\\
	&&	
	\hspace{2cm}
	\cdots R_{\ell+1}^T H^{(\ell+1)}(\tau^{(\ell,1)}(x)) R_{\ell+1}
	\partial_{(R_\ell)_{jk}}\tau^{(\ell)}
	\big(\tau^{(\ell-1,1)}(x)\big)
	\nonumber\\
	&=&P^+_{\ell+1,Q}\partial_{(R_\ell)_{jk}}\tau^{(\ell)}
	\big(\tau^{(\ell-1,1)}(x)\big) \,.
\eeqn 
Therefore,
\eqn 
	\lefteqn{
	\partial_{(R_\ell)_{jk}}
	\frac12|\tau^{(Q,1)}(x)-\widetilde y_{\ell'}|^2
	}
	\nonumber\\
	&=&
	\sum_{m,i}(P^+_{\ell+1,Q})_{mi}
	(\partial_{(R_\ell)_{jk}}\tau^{(\ell)}
	\big(\tau^{(\ell-1,1)}(x)\big) )_i
	(\tau^{(Q,1)}(x)-\widetilde y_{\ell'})_m
	\nonumber\\
	&=&
	\sum_{m,i}(P^+_{\ell+1,Q})_{mi}
	(\tau^{(Q,1)}(x)-\widetilde y_{\ell'})_m 
	\Big(\delta_{ki}
	\sigma(R_{\ell}(\tau^{(\ell-1,1)}(x)+\beta^{(\ell)}))_j
	\nonumber\\
	&&\hspace{1cm}
	+(R_\ell)_{ji}h(R_{\ell}(\tau^{(\ell-1,1)}(x)+\beta^{(\ell)}))_j
	(\tau^{(\ell-1,1)}(x)+\beta^{(\ell)})_k\Big)  
	\nonumber\\
	&=&
	\sigma(R_{\ell}(\tau^{(\ell-1,1)}(x)+\beta^{(\ell)}))_j
	\sum_{m}(P^+_{\ell+1,Q})_{mk} 
	(\tau^{(Q,1)}(x)-\widetilde y_{\ell'})_m
	\nonumber\\
	&& 
	+h(R_{\ell}(\tau^{(\ell-1,1)}(x)+\beta^{(\ell)}))_j\sum_{m,i}
	(R_\ell)_{ji}
	(P^+_{\ell+1,Q})_{mi}(\tau^{(Q,1)}(x)-\widetilde y_{\ell'})_m
	\nonumber\\
	&&\hspace{2cm}
	(\tau^{(\ell-1,1)}(x)+\beta^{(\ell)})_k \,,
\eeqn 
and in matrix notation,
\eqn 
	\lefteqn{
	[\partial_{(R_\ell)_{jk}}
	\frac12|\tau^{(Q,1)}(x)-\widetilde y_{\ell'}|^2]
	}
	\nonumber\\
	&=&
	\sigma(R_{\ell}(\tau^{(\ell-1,1)}(x)+\beta^{(\ell)}))
	(\tau^{(Q,1)}(x)-\widetilde y_{\ell'})^T P^+_{\ell+1,Q}
	\\
	&&\hspace{1cm} 
	+H^{(\ell)} 
	R_{\ell}
	(P^+_{\ell+1,Q})^T(\tau^{(Q,1)}(x)-\widetilde y_{\ell'})
	(\tau^{(\ell-1,1)}(x)+\beta^{(\ell)})^T \,,
	\nonumber\\
	&=&
	H^{(\ell)} R_{\ell}(\tau^{(\ell-1,1)}(x)+\beta^{(\ell)})
	((P^+_{\ell+1,Q})^T(\tau^{(Q,1)}(x)-\widetilde y_{\ell'}))^T 
	\\
	&&\hspace{1cm} 
	+H^{(\ell)} 
	R_{\ell}
	(P^+_{\ell+1,Q})^T(\tau^{(Q,1)}(x)-\widetilde y_{\ell'})
	(\tau^{(\ell-1,1)}(x)+\beta^{(\ell)})^T \,,
	\nonumber
\eeqn 
using $\sigma(x)=H(x)x$.
\endprf

Therefore,
\eqn
	\widetilde\Omega_\ell &=& 
	-\pi_-\Big(\sum_{\ell'}\int dx\,\mu_{\ell'}(x)
	(\partial_{R_\ell}
	\frac12|\tau^{(Q,1)}(x)-\widetilde y_{\ell'}|^2)\Big)R_\ell^T\Big)
	\nonumber\\
	&=&
	-\sum_{\ell'}\int dx\,\mu_{\ell'}(x)[H^{(\ell)} ,
	M_{\ell,\ell'}(x)] 
\eeqn 
where
\eqn
	M_{\ell,\ell'}(x) &=& \frac12 R_{\ell}\Big(
	(P^+_{\ell+1,Q})^T(\tau^{(Q,1)}(x)-\widetilde y_{\ell'})
	(\tau^{(\ell-1,1)}(x)+\beta^{(\ell)})^T  
	\\
	&&\hspace{1cm}
	+  (\tau^{(\ell-1,1)}(x)+\beta^{(\ell)})
	((P^+_{\ell+1,Q})^T(\tau^{(Q,1)}(x)-\widetilde y_{\ell'}))^T \Big)
	R_\ell^T \,,
	\nonumber
\eeqn 
as claimed.
\qed


$\;$\\
\noindent
{\bf Acknowledgments:} 
The author thanks Patricia Mu\~{n}oz Ewald, Andrew Moore, and Vardan Papyan for discussions, and gratefully acknowledges support by the NSF through the grant DMS-2009800, and the RTG Grant DMS-1840314 - {\em Analysis of PDE}. 
\\

\end{document}